\documentclass{article}

\usepackage[round]{natbib}

\usepackage[utf8]{inputenc} % allow utf-8 input
\usepackage[T1]{fontenc}    % use 8-bit T1 fonts
\usepackage{lmodern}
\usepackage{hyperref}       % hyperlinks
\usepackage{url}            % simple URL typesetting
\usepackage{booktabs}       % professional-quality tables
\usepackage{amsfonts}       % blackboard math symbols
\usepackage{nicefrac}       % compact symbols for 1/2, etc.
\usepackage{microtype}      % microtypography
\usepackage{xcolor}         % colors
\usepackage[pdftex]{graphicx}
\usepackage{subcaption}
\usepackage{mathtools}
\usepackage{algorithm}
\usepackage{algorithmic}
\usepackage{makecell}
\usepackage{multirow}
\usepackage{longtable}
\usepackage{adjustbox}
\usepackage{wrapfig}
\usepackage{makecell}
\title{Improving Deep Ensembles by Estimating Confusion Matrices}

\author{Danil Kuzin \\ Lancaster University \and Olga Isupova \\ University of Oxford \and  Steven Reece \\ University of Oxford \and Brooke D Simmons \\ Lancaster University}
\date{17 Oct 2024}
\begin{document}
\maketitle

\begin{abstract}
  Ensembling in deep learning improves accuracy and calibration over single networks. The traditional aggregation approach, ensemble averaging, treats all individual networks equally by averaging their outputs. Inspired by crowdsourcing we propose an aggregation method called soft Dawid Skene for deep ensembles that estimates confusion matrices of ensemble members and weighs them according to their inferred performance. Soft Dawid Skene aggregates soft labels in contrast to hard labels often used in crowdsourcing. We empirically show the superiority of soft Dawid Skene in accuracy, calibration and out of distribution detection in comparison to ensemble averaging in extensive experiments.
\end{abstract}

\section{Introduction}
\label{sec:intro}
Ensembling is a popular approach to improve accuracy and calibration of deep neural networks~\citep{lakshminarayanan2016simple}.  Predictions of individual neural networks are aggregated by averaging predicted probabilities, this is called ensemble averaging (EA). Despite its simplicity, EA shows strong results in accuracy, calibration and uncertainty estimation outperforming other Bayesian and non-Bayesian specialised methods~\citep{ashukha2020pitfalls, ovadia2019can, wortsman2022model, pei2022transformer, kim2023unified, gustafsson2020evaluating, franchi2022muad}.

However, this simplest approach does not take into account the structural differences of the ensemble members.  Consider an ensemble of three neural networks.  Model A specializes in recognizing animals but frequently misclassifies vehicles. Model B is the opposite --- highly accurate at vehicles but poorer at animals. Model C has the average performance on both.  EA gives same weights for all three models when making predictions, but model A should be trusted more for animals, while model B should be preferred for vehicles, and model C should complement their predictions.  Ideally, we want an aggregation method to infer individual strenghts and weaknesses% of each model% based on their predictions for different classes.  It should
and then weigh models accordingly, rather than treating all models equally.

One way to estimate the expertise of ensemble members is through confusion matrices, which show how often a model predicts each class versus the true class.  In the example above, model C has a diagonal confusion matrix, the confusion matrix for model A has higher diagonal weights for animal rows, but more uniform weights for vehicle rows.  Confusion matrices quantify the expertise across ensemble members. However, we do not have access to ground truth classes while making predictions, thus the aggregation method should estimate confusion matrices.

%The structural differences are usually present even in the models of the same architecture trained on the full dataset.  These models usually have slightly different confusion matrices, and weighing their predictions equally, as it is done by EA, is suboptimal.  In cases when models are trained on the splits of the training data (e.g., bagging) or are of different quality, proper weighing of the predictions may provide significant improvement.  Therefore, a selective ensemble technique should analyse these confusion matrices to infer the reliability of each model on different classes. The predictions can then be aggregated, weighting more towards members that are expected to be more accurate.

\begin{figure*}[tb]
    \centering
    \begin{subfigure}[t]{0.49\textwidth}
        \centering
        \includegraphics[width=1\linewidth]{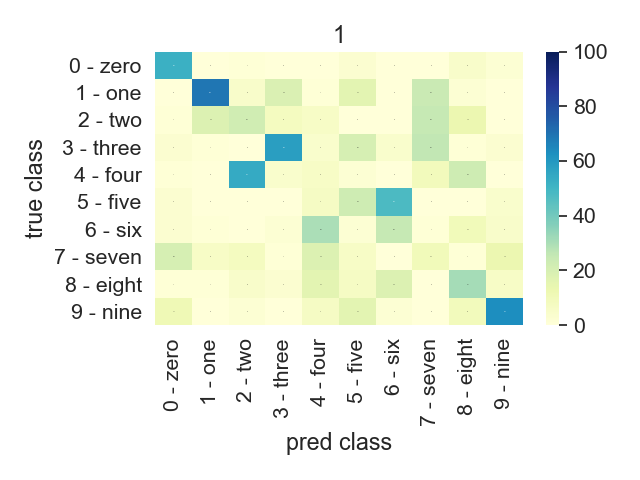}
        % \caption{accuracy (\(\uparrow\))}
        \label{motivation:cm1}
    \end{subfigure}
    \begin{subfigure}[t]{0.49\textwidth}
        \centering
        \includegraphics[width=1\linewidth]{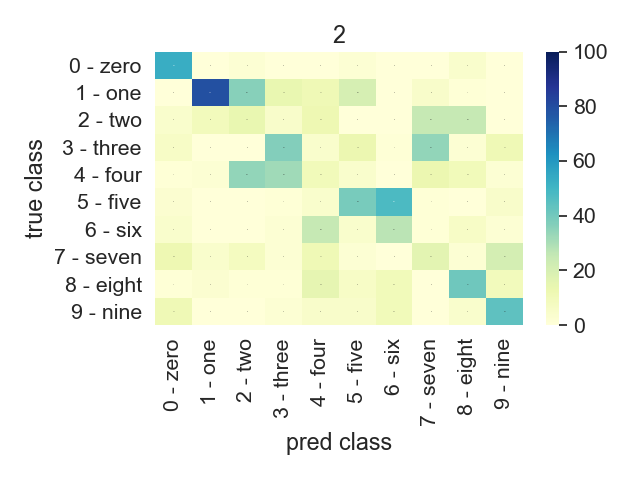}
        % \caption{ECE (\(\downarrow\))}
        \label{motivation:cm2}
    \end{subfigure}
    \caption{True confusion matrices for two ensemble members on the MNIST distributional shift experiment with rotation of $60^{\circ}$.}
    \label{motivation:mnist_cms}
\end{figure*}

Aggregation of different classifier predictions and estimation of their confusion matrices is explored in crowdsourcing literature~\citep{sheng2019machine}.
%As usually inputs from volunteers do not provide probabilities, only hard labels, these methods operate on hard labels. In the majority voting algorithm the mode of the categories from the multiple classifiers is chosen.  If there are ties, the category is chosen randomly.  In cases with categories imbalance, weights for categories can be added to the model.  Aggregated confidence of classifiers can be computed as fraction of votes for each category.
Majority voting (MV) is a baseline in which the mode of class predictions is chosen. MV weighs each classifier equally and does not estimate confusion matrices (similar to EA). The popular approach that estimates confusion matrices and hence weighs the classifiers differently is Dawid Skene (DS)~\citep{dawid1979maximum}, which uses the expectation-maximization (EM) algorithm~\citep{dempster1977maximum}.  The Bayesian version of this model is Bayesian classifier combination (BCC)~\citep{ghahramani2003bayesian, simpson2013dynamic}.%, is usually implemented with Markov chain Monte Carlo sampling~\citep{ghahramani2003bayesian} or variational inference~\citep{simpson2013dynamic} algorithms.

Crowdsourcing is usually used to aggregate labels from humans, who normally provide only a class label for each data point without confidence of this label, i.e. a \emph{hard label}. When aggregating the outputs of neural networks the output probabilities are usually available, i.e. \emph{soft labels}.  These outputs can be biased or  miscalibrated~\citep{guo2017calibration}. Despite this miscalibration, it has been demonstrated, soft labels improve the generalisation of neural networks~\citep{hinton2015distilling, peterson2019human, uma2020case, grossmann2022beyond}. Therefore, we propose to use soft labels for aggregating predictions of an ensemble.

In the context of crowdsourcing, EA corresponds to MV, where outputs of each crowd member have an equal weight in the consensus answer. MV has been a strong baseline in crowdsourcing literature (as EA in ensembling) and is often used in practice due to its simplicity. However, for hard labeling, crowdsourcing models that weigh crowd members differently according to their skills outperform MV~\citep{dawid1979maximum, ghahramani2003bayesian, zhou2012learning, simpson2013dynamic, li2019exploiting}. Using this insight, we explore whether using aggregation with different weights would be beneficial for soft labels as well.

We propose an extension of DS model to soft labels, which we call Soft Dawid Skene (SDS). DS assumes the crowd labels have the multinomial distribution which we replace with the Dirichlet distribution.  We still use the EM-algorithm, but M step is no longer analytically tractable, so we use the automatic differentiation approach~\citep{paszke2019pytorch}.  We also add Polyak averaging to the E step to stabilize the convergence~\citep{polyak1990automat}.  We then conduct extensive experiments on MNIST, CIFAR10/100, and ImageNet distributional shift data, showing that estimation of the confusion matrices improves the accuracy and calibration of the ensemble. We also conduct the out-of-distribution detection experiments to demonstrate usefulness of inferred uncertainties. Our main contributions are:
\begin{itemize}
    \item We propose a soft label voting model and algorithm for aggregation of classifiers.
    \item We apply this algorithm to aggregate predictions of an ensemble of neural networks.
    \item Experiments on distributional shift data and out-of-distribution detection show superiority of the proposed algorithm to baselines.
\end{itemize}

\section{Motivation}

\begin{figure*}[tb]
    \centering
    \begin{subfigure}[t]{0.48\textwidth}
        \centering
        \includegraphics[scale=0.8]{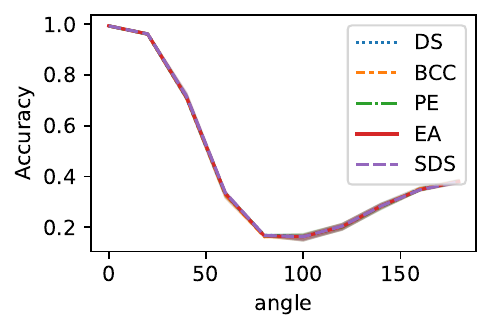}
        \caption{accuracy (\(\uparrow\))}
        \label{experiments:mnist_acc}
    \end{subfigure}
    \begin{subfigure}[t]{0.48\textwidth}
        \centering
        \includegraphics[scale=0.8]{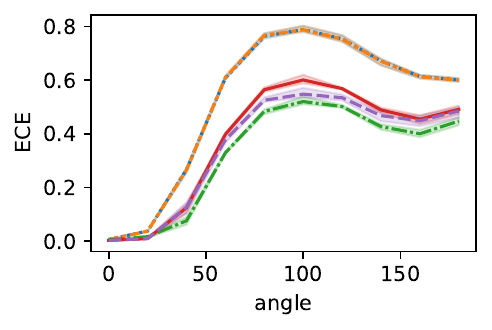}
        \caption{ECE (\(\downarrow\))}
        \label{experiments:mnist_ece}
    \end{subfigure}
    \hfill
    \begin{subfigure}[t]{0.48\textwidth}
        \centering
        \includegraphics[scale=0.8]{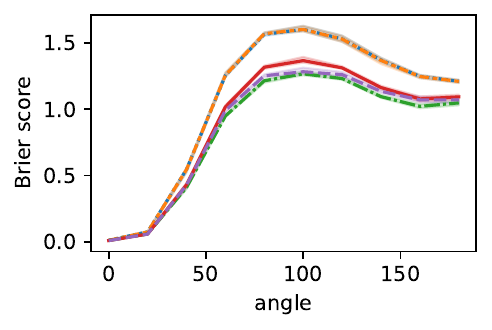}
        \caption{Brier Score (\(\downarrow\))}
        \label{experiments:mnist_brier}
    \end{subfigure}
    \begin{subfigure}[t]{0.48\textwidth}
        \centering
        \includegraphics[scale=0.8]{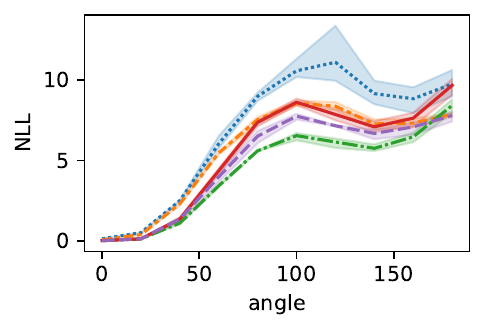}
        \caption{NLL (\(\downarrow\))}
        \label{experiments:mnist_nll}
    \end{subfigure}
    \caption{Results on MNIST with rotations of increasing angle: (\subref{experiments:mnist_acc}) accuracy, (\subref{experiments:mnist_ece}) ECE, (\subref{experiments:mnist_brier}) Brier score, (\subref{experiments:mnist_nll}) NLL. Here, several methods demonstrate similar performance, and some lines are not visible. Thus, all methods have similar accuracy and DS and BCC methods have similar ECE and Brier score.}
    \label{experiments:mnist}
\end{figure*}

% \subsection{Confusion Matrices Estimation}
Even for simple identical architectures and small datasets, ensemble members can have different expertise, therefore weighing them equally for the aggregated prediction, as it is done in EA, may be suboptimal.

To demonstrate it, \figurename~\ref{motivation:mnist_cms} presents the confusion matrices for two different ensemble members that show the proportion of predicted and true labels for each class.  These are results for the MNIST dataset~\citep{lecun1998gradient} with additionally applied rotation transformation for validation data (a detailed explanation of the experimental setup is given in section~\ref{sec:experiments}). Although these confusion matrices are generated for the ensemble members trained in the same way, there are multiple substantial differences in them. For example, when model \(2\) predicts ``3'' the probabilities that the true class for this data point are ``3'' or ``4'' are almost equal. In contrast, model \(1\) predicts ``3'' with the high probability of being correct. Therefore, when we combine predictions of these models, we should weigh model \(1\) more than model \(2\) when they predict ``3''.  This example shows that the confusion matrices contain valuable information about reliability of the ensemble members, which can be used for inferring joint predictions.

%confusion matrices of ensemble members can be different, especially under the distributional shift.
%Even for simple architectures and small datasets confusion matrices of ensemble members can be different, especially under the distributional shift.

%We can expect to see larger differences in confusion matrices, and thus reliability, of different ensemble members for more complex datasets and when ensemble members have different modelling capacity, e.g, have different architectures and are trained on slightly different training sets.

%different values in them and thus estimation of these matrices can add structural information that improves predictions, as it is shown in experimental section.  Moreover, this is the simple example and for more complex scenarios the difference is larger.

% \subsection{Ensembling with Soft Labels}

% The previous example with confusion matrices suggests using them while weighing ensemble member predictions.
The confusion matrices in \figurename~\ref{motivation:mnist_cms} are computed using the ground truth labels, which are not available in practice. There are methods from crowdsourcing literature that estimate these confusion matrices on the fly without access to the ground truth, such as~\citep{dawid1979maximum, simpson2013dynamic}. However, most of them use only hard labels, that are class assignments from crowd members.
Therefore, they ignore all the information about individual model uncertainty encoded in soft labels, i.e., predicted probabilities of each class, when they are applied in ensembling.
%Therefore, they cannot be directly applied for the case of soft labels in ensembling, as they do not use uncertainty information from individual predicted probabilities of each model.

As it turns out this information encoded in soft labels is important for inferring ensemble predictions, because EA, that does not use confusion matrices but uses soft labels, outperforms crowdsourcing models, that estimate confusion matrices but use hard labels. This is despite the fact that probability estimates of modern neural networks can be miscalibrated. \figurename~\ref{experiments:mnist} shows this on the rotated MNIST example (more details about the experiment and metrics used can be found in section~\ref{sec:experiments}).  Unsurprisingly, here crowdsourcing models DS~\citep{dawid1979maximum} and BCC~\citep{simpson2013dynamic}, that use hard labels and thus ignore ensemble members' uncertainty, show worse results in calibration metrics (expected calibration error (ECE)~\citep{naeini2015obtaining} and Brier score~\citep{brier1950verification}) than EA~\citep{lakshminarayanan2016simple}, that utilises soft labels.

In these two examples in \figurename~\ref{motivation:mnist_cms} and~\ref{experiments:mnist} we have shown that both model reliability, that can be estimated with a confusion matrix, and model uncertainty encoded in soft labels, are useful for inferring accurate and calibrated predictions of an ensemble. Therefore, we propose the {\it Soft Dawid Skene model} that uses both these sources of information.

\section{Soft Dawid Skene (SDS) model}
\label{sec:sds_model}
% Notation:
% \(N\) --- size of dataset.
% \(J\) --- number of classes.
% \(K\) --- number of classifiers.

We assume that there is a dataset of \(N\) data points and there are \(K\) classifiers.  Data points belong to one of the \(J\) true labels \(t_i \in \{1, ..., J\}, i \in \{1,\ldots,N\}\). However, we only observe probabilities of each of \(J\) labels predicted by classifiers. Let \(\mathbf{c}_i^{(k)}=\{c^{(k)}_{il}\}_{l=1}^J\in [0, 1]^J\) denote the vector of probabilities of classifier \(k\) for data item \(i\), \(k \in \{1,\ldots,K\}\), \(i \in \{1,\ldots,N\}\).  We assume that classifiers are conditionally independent given the true labels and each has its own confusion matrix \(\boldsymbol\Pi^{(k)}=\{\pi_{jl}^{(k)}\}_{j, l=1}^J\in [0, \infty)^{J\times J}\), which is a prior matrix of probabilities of misclassification, \(\pi^{(k)}_{jl}\) is the indicator of the probability that classifier \(k\) predicts label \(l\) when the true label is \(j\). By an `indicator' here we mean the parameter of the corresponding Dirichlet distribution as it is shown below. Hereafter, we will refer to \(\boldsymbol\Pi^{(k)}\) as confusion matrices for simplicity.

The assumption of classifier conditional independence is strong as predictions of neural networks may be correlated.  However, we show empirically that even with this assumption predictions made by our model are better than EA.  Some work on correlated confusion matrices in crowdsourcing is undertaken in, e.g.,~\citep{ghahramani2003bayesian}, but it is out of scope of this paper.

% Modelling
We model the distribution for \(t_i\) as multinomial with parameters \(\boldsymbol\nu = \{\nu_j\}_{j=1}^J\in [0, 1]^J\):\(    p(t_i=j|\boldsymbol\nu)=\nu_j\).

The prior of classifiers' predicted probabilities \(\mathbf{c}_i^k\) is set as the Dirichlet distribution:  \(\mathbf{c}_i^{(k)} \sim \text{Dir}(\boldsymbol\pi^{(k)}_{t_i})\), where \(\boldsymbol\pi^{(k)}_{t_i} = \{\pi^{(k)}_{t_i l}\}_{l=1}^J\). That is
%\begin{equation}
\[
    p(\mathbf{c}_i^{(k)}|\boldsymbol\pi^{(k)}_{t_i}) = \frac{1}{B(\boldsymbol\pi^{(k)}_{t_i})}\prod_{l=1}^J {c_{il}^{(k)}}^{\pi_{t_i l}^{(k)}-1},
\]
%\end{equation}
where \(B(\cdot)\) is the Beta function. And with the assumption of conditional independence of classifiers, the distribution of predictions of all classifiers \(\mathbf{c}_i = \{\mathbf{c}_i^{(k)}\}_{k=1}^{K}\) is
%\begin{equation}
\(
    p(\mathbf{c}_i | t_i, \boldsymbol\Pi) = \prod_k p(\mathbf{c}_i^{(k)}|\boldsymbol\pi^{(k)}_{t_i}),
\)
%\end{equation}
where \(\boldsymbol\Pi = \{\boldsymbol\Pi^{(k)}\}_{k=1}^K\).

To summarise, in SDS the observed data is classifiers' labels \(\mathbf{C} = \{\mathbf{c}_i\}_{i=1}^N\), latent data is true labels \(\mathbf{t} = \{t_i\}_{i=1}^N\), and parameters are priors of class probabilities \(\boldsymbol\nu \) and confusion matrices \(\boldsymbol\Pi\).

\section{EM algorithm}
\label{sec:em}
An EM algorithm~\citep{dempster1977maximum} is an iterative algorithm that fits the model by finding the maximum likelihood estimate of the marginal likelihood
%\begin{equation}
\(
    p(\mathbf{C} \mid \boldsymbol\nu, \boldsymbol\Pi)=
    \int p(\mathbf {C}, \mathbf {t} \mid \boldsymbol\nu, \boldsymbol\Pi)\,d\mathbf {t}
    % \int p(\mathbf {X} ,\mathbf {Z} \mid {\boldsymbol {\theta }})\,d\mathbf {Z} =\int p(\mathbf {X} \mid \mathbf {Z} ,{\boldsymbol {\theta }})p(\mathbf {Z} \mid {\boldsymbol {\theta }})\,d\mathbf {Z} }
\)
%\end{equation}
using the \(Q\) function:
%\begin{equation}
\[
    Q(\boldsymbol\nu, \boldsymbol\Pi\mid \boldsymbol\nu^{\text{old}}, \boldsymbol\Pi^{\text{old}})=\text{E}_{\mathbf {t} \mid \mathbf {C} ,\boldsymbol\nu^{\text{old}}, \boldsymbol\Pi^{\text{old}}}\left[\log p(\mathbf {C}, \mathbf {t} \mid \boldsymbol\nu, \boldsymbol\Pi)\right],
\]
%\end{equation}
where \(\boldsymbol\nu^{\text{old}}, \boldsymbol\Pi^{\text{old}}\) are the values of parameters at the previous iteration.
In our model the \(Q\) function is
\begin{equation}
    \label{eq:Q}
    \begin{split}
        Q =& \sum_i \sum_j p(t_i=j \mid \mathbf{C}, \boldsymbol\nu^{\text{old}}, \boldsymbol\Pi^{\text{old}}) \\
        &\times \biggl[\log\nu_j + \sum_k \sum_l (\pi_{j l}^{(k)}-1) \log c_{il}^{(k)} \\
        &- \sum_k \Bigl( \sum_l \log \Gamma(\pi_{j l}^{(k)}) - \log \Gamma\big(\sum_l \pi_{j l}^{(k)}\big)\Bigr)\biggr],
    \end{split}
\end{equation}
where \(\Gamma(\cdot)\) is the Gamma function.
And we want to find the parameters that maximize it by iterating E and M steps:
\(Q \to \max_{\boldsymbol\Pi,\boldsymbol\nu}\).

\paragraph{M Step.}
In the maximization (M) step, we find the new estimates of the parameters that maximize~\(Q\):
%\begin{equation}
\(
    \boldsymbol\nu^{\text{new}}, \boldsymbol\Pi^{\text{new}}={\operatorname {arg\,max} }_{\boldsymbol\nu, \boldsymbol\Pi}\ Q(\boldsymbol\nu, \boldsymbol\Pi \mid \boldsymbol\nu^{\text{old}}, \boldsymbol\Pi^{\text{old}}).
\)
%\end{equation}
Hereafter, we omit the superscripts ``new'' and ``old'' for readability.

\textit{Maximise \(\boldsymbol\nu\).}  Maximization of \(\boldsymbol\nu \) subject to \(\sum_j \nu_j = 1\), as this is a probability distribution, is a convex constrained optimization problem, which can be solved with Lagrange multipliers:
%\begin{equation}
\(
    % \begin{split}
        %        \mathcal{L}_{\boldsymbol\nu} =& \sum_i \sum_j p(t_i=j \mid \mathbf{C}, \boldsymbol\nu^{\text{old}},  \boldsymbol\Pi^{\text{old}}) \log\nu_j \\
        %        &- \lambda(\sum_j \nu_j - 1) \to \max_{\boldsymbol\nu}
        \mathcal{L}_{\boldsymbol\nu} = \sum_i \sum_j p(t_i=j \mid \mathbf{C}, \boldsymbol\nu,  \boldsymbol\Pi) \log\nu_j - \lambda(\sum_j \nu_j - 1) \to \max_{\boldsymbol\nu},
    % \end{split}
\)
%\end{equation}
where \(\lambda\) is a Lagrange multiplier. The optimal class priors is just a mean probability of class being \(j\) given the current estimates of confusion matrices and observed classifier labels:
\begin{equation}
    \label{eq:m_step_nu}
    % \nu_j = \frac{\sum_i p(t_i=j | \mathbf{C}, \boldsymbol\nu^{\text{old}},  \boldsymbol\Pi^{\text{old}})}{\sum_i \sum_{j'} p(t_i=j' | \mathbf{C}, \boldsymbol\nu^{\text{old}}, \boldsymbol\Pi^{\text{old}})}
    \nu_j = \frac{\sum_i p(t_i=j | \mathbf{C}, \boldsymbol\nu,  \boldsymbol\Pi)}{\sum_i \sum_{j'} p(t_i=j' | \mathbf{C}, \boldsymbol\nu, \boldsymbol\Pi)}.
\end{equation}

\textit{Maximise \(\boldsymbol\Pi\).}  In this step we want to find the new estimates of confusion matrices \(\boldsymbol\Pi\) that maximize~\(Q\).  In our model confusion matrices are not required to be normalized, as they are only used as priors for the Dirichlet distributions.  %Therefore, we can maximize \(Q\) w.r.t. \(\boldsymbol\Pi\) without any constraints.
However, the gradient of \(Q\) with respect to \(\boldsymbol\Pi\) involves the digamma functions, and the updated values of \(\boldsymbol\Pi\) are analytically intractable.

Instead, we use automatic differentiation from PyTorch~\citep{paszke2019pytorch} to perform the M step for \(\boldsymbol\Pi \). That is, we set \(Q\) function (eq.~(\ref{eq:Q})) as a loss function and \(\boldsymbol\Pi\) as parameters for optimisation. Additionally, optimiser's weight decay limits the large values, that would otherwise cause instability at the E step.  We use AdamW optimiser~\citep{loshchilov2017decoupled}% with the grid-optimised learning rate
. The exact values of weight decay play a small role, just some weight decay needs to be present.  We do not need to run the optimiser until convergence, as we have an outer loop of the EM algorithm. We found that \(5\) optimiser steps are sufficient.

\paragraph{E Step.}
At the expectation (E) step we compute the expectations of the \(Q\) function to update estimates of latent \(\mathbf{t}\) with the current estimates of \(\boldsymbol\nu\) and \(\boldsymbol{\Pi}\):
\begin{align}
    \label{eq:e_step}
    %\begin{split}
        % \log \,&p(t_i=j | \mathbf{C}, \boldsymbol\nu^{\text{old}}, \boldsymbol\Pi^{\text{old}}) \propto  \log\nu_j \\
        % & + \sum_k \sum_l (\pi_{j l}^{(k)}-1) \log c_{il}^{(k)}\\
        % & - \sum_k \left( \sum_l \log \Gamma(\pi_{j l}^{(k)}) - \log \Gamma(\sum_l \pi_{j l}^{(k)})\right),
        % \log \,&p(t_i=j | \mathbf{C}, \boldsymbol\nu, \boldsymbol\Pi) \propto  \log\nu_j + \sum_{k, l} (\pi_{j l}^{(k)}-1) \log c_{il}^{(k)}\\
        % & - \sum_k \left( \sum_l \log \Gamma(\pi_{j l}^{(k)}) - \log \Gamma(\sum_l \pi_{j l}^{(k)})\right).
    %\end{split}
    %\begin{align}
        &\log p(t_i=j | \mathbf{C}, \boldsymbol\nu, \boldsymbol\Pi) \propto  \log\nu_j + \sum_{k, l} (\pi_{j l}^{(k)}-1) \log c_{il}^{(k)} \nonumber\\
        & - \sum_k \left( \sum_l \log \Gamma(\pi_{j l}^{(k)}) - \log \Gamma(\sum_l \pi_{j l}^{(k)})\right).
    %\end{align}
\end{align}

%(Here we omit the superscript "old" for readability.)

Let \(p_{\text{new}}(t_i)\) be the new estimates of \(\mathbf{t}\) from eq.~(\ref{eq:e_step}), \(p_{\text{old}}(t_i)\) be the current estimates of \(\mathbf{t}\), and \(\alpha \in [0, 1]\) be the weight of the new estimate. We use Polyak averaging~\citep{polyak1990automat}:
\begin{equation}
    \label{eq:polyak}
    % p(t_i = j | \mathbf{C}, \boldsymbol{\nu}^{\text{old}}, \boldsymbol{\Pi}^{\text{old}}) = (1 - \alpha)p_{\text{old}}(t_i) + \alpha \, p_{\text{new}}(t_i),
    p(t_i = j | \mathbf{C}, \boldsymbol{\nu}, \boldsymbol{\Pi}) = (1 - \alpha)p_{\text{old}}(t_i) + \alpha \, p_{\text{new}}(t_i),
\end{equation}
%where \(p_{\text{new}}(t_i)\) is the new estimates of \(\mathbf{t}\) from eq.~(\ref{eq:e_step}), \(p_{\text{old}}(t_i)\) is the current estimates of \(\mathbf{t}\), and \(\alpha \in [0, 1]\) is the weight of the new estimate.
%that is a weighted average of new and previous estimates of \(\mathbf{t}\) to improve the stability of the algorithm due to the approximate M step.
to improve the stability of the algorithm due to the approximate M step.

\subsection{Full Algorithm}
The EM algorithm requires initial values of the parameters and latent variables and stopping criteria.  For initialization of \(\boldsymbol\Pi\) and \(\boldsymbol\nu\), we use one iteration of the DS algorithm~\citep{dawid1979maximum} on hard versions of \(\mathbf{C}\), i.e., class values that have the maximum value in the corresponding probability vector.  For the initial values of probabilities of \(\mathbf{t}\) we use the EA algorithm~\citep{lakshminarayanan2016simple}.  The EM algorithm iterates until convergence of the \(Q\) function. However, in practice we found that the fixed number of iterations is sufficient. The full algorithm is shown in Algorithm \ref{alg:em_for_sds_alg}. The values of all hyperparameters of the algorithm used in the experiments can be found in Appendix D. We provide an illustrative example how SDS works in Appendix C. %~\ref{sec:hyperparameters}.

% \begin{wrapfigure}{r}{0.5\textwidth}
% \vspace{-25px}
% \begin{minipage}{0.49\textwidth}
\begin{algorithm}[H]
    \caption{EM algorithm for SDS}\label{alg:em_for_sds_alg}
    \begin{algorithmic}
        \STATE Initialise \(\boldsymbol\Pi\) and \(\boldsymbol\nu\) with the DS algorithm
        \STATE Initialise \(\mathbf{t}\) with the EA algorithm
        \WHILE{true}
        \STATE \quad \textbf{E step:}
        \STATE \quad \quad Compute \(\log p(t_i=j | \mathbf{C}, \boldsymbol\Pi^{\text{old}})\) for all \(i, j\) (\ref{eq:e_step})
        \STATE \quad \quad Update \(\mathbf{t}\) with Polyak averaging (\ref{eq:polyak})
        \STATE \quad \textbf{M step:}
        \STATE \quad \quad Update \(\boldsymbol\nu\) using (\ref{eq:m_step_nu})
        \STATE \quad \quad Update \(\boldsymbol\Pi\) with the automatic differentiation optimiser on (\ref{eq:Q})
        \STATE Check convergence of \(Q\) function
        \ENDWHILE
    \end{algorithmic}
\end{algorithm}
% \end{minipage}
% \end{wrapfigure}

\section{Related works}
Ensemble learning and crowdsourcing are two relevant areas for weighted aggregation of models.

\textit{Ensembling.}
Despite the popularity of ensembling only EA or hard label MV are used for aggregation. Many works on ensembling focus on diversifying ensemble members: e.g., bagging~\citep{breiman1996bagging, huang2015improving, rew2021robust, wei2024bend}, particle-based inference,~\citep{d2021repulsive}, or others~\citep{jain2020maximizing, zhang2020diversified}. These approaches are complimentary to ours as they still use averaging for aggregation of ensemble member predictions.

We consider the case where pre-trained neural networks are used in an ensemble to predict on new data, without access to the training data during prediction and ensembling. This is in contrast to boosting-type and stacked ensembling methods, when a top-level classifier is trained using responses from ensemble members, e.g.~\citep{mao1998case, zhang2020snapshot, moghimi2016boosted, young2018deep, das2022design, li2022novel, walach2016learning, li2022trustworthy}. This is also different to probabilistic ensembling~\citep{wang2023diversity}, which builds a mixture of Gaussians (MoG) using Laplace approximation~\citep{daxberger2021laplace} and requires training labels. Ensemble members are trained sequentially in a boosting manner.

To the best of our knowledge, from the ensembling literature~\citep{ganaie2022ensemble, dong2020survey} EA or hard label MV are the only methods for prediction aggregation that exist for our setting. Averaging method from~\citep{huang2015improving} can satisfy our setting but it is only applicable for the autoencoder neural networks.

\begin{figure*}[!t]
    \centering
    \begin{subfigure}[t]{0.49\textwidth}
        \centering
        \includegraphics[scale=0.8]{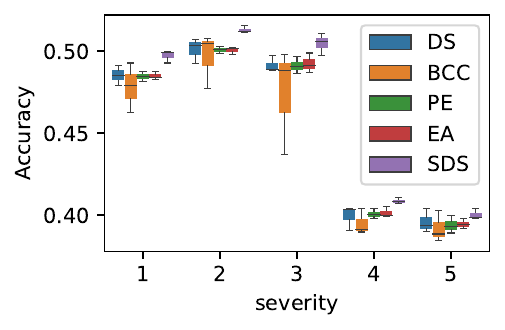}
        \caption{accuracy (\(\uparrow\))}
        \label{experiments:corrupted_cifar10_acc}
    \end{subfigure}
    \begin{subfigure}[t]{0.49\textwidth}
        \centering
        \includegraphics[scale=0.8]{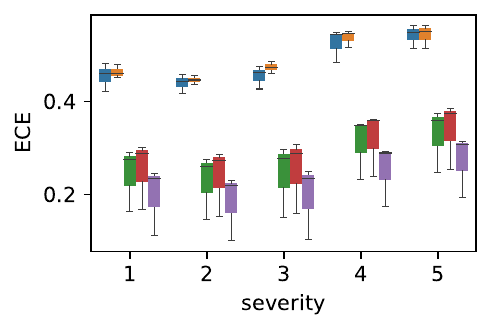}
        \caption{ECE (\(\downarrow\))}
        \label{experiments:corrupted_cifar10_ece}
    \end{subfigure}
    \hfill
    \begin{subfigure}[t]{0.49\textwidth}
        \centering
        \includegraphics[scale=0.8]{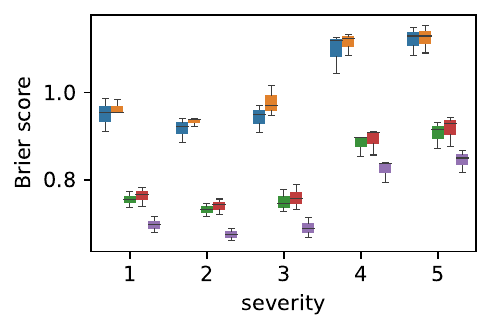}
        \caption{Brier Score (\(\downarrow\))}
        \label{experiments:corrupted_cifar10_brier}
    \end{subfigure}
    \begin{subfigure}[t]{0.49\textwidth}
        \centering
        \includegraphics[scale=0.8]{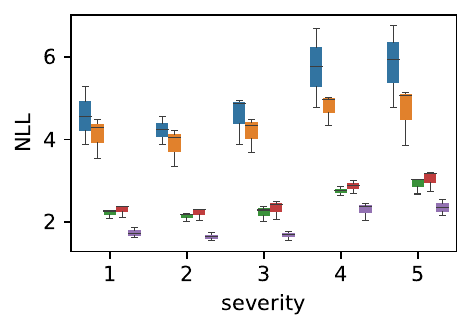}
        \caption{NLL (\(\downarrow\))}
        \label{experiments:corrupted_cifar10_nll}
    \end{subfigure}
    \caption{Results on CIFAR10 with Frosted Glass Blur corruption of increasing severity: (\subref{experiments:corrupted_cifar10_acc}) accuracy, (\subref{experiments:corrupted_cifar10_ece}) ECE, (\subref{experiments:corrupted_cifar10_brier}) Brier score, (\subref{experiments:corrupted_cifar10_nll}) NLL.}
    \label{experiments:corrupted_cifar10}
\end{figure*}

\textit{Crowdsourcing.}
While the majority of the crowdsourcing literature use hard labels, there are a few works that deal with soft labels from crowd members~\citep{nazabal2015human, mendez2022eliciting, augustin2017bayesian, chung2019efficient}. However, to the best of our knowledge none are applicable to our setting of ensembling deep learning models. They either use the mode (or median) of soft crowd labels~\citep{mendez2022eliciting, chung2019efficient, collins2022eliciting}, i.e., the equivalent of EA, or they are used for the problems where the ground truth labels are expected to be soft~\citep{augustin2017bayesian}, e.g., in text categorization where a label for each document is the set of proportions of each category in the text. In contrast to the latter setting we expect soft labels from ensemble (crowd) members but the ground truth label to be a single class label, i.e., a hard label.

\citet{nazabal2015human} similarly to us propose a soft label extension of the existing crowdsourcing method, BCC in their case. \citet{steyvers2022bayesian} at the intersection of crowdsourcing and ensembling combine soft predictions from human and machine classifiers. However, in contrast to our method, inference of test ground truth classes in both these works requires at least part of the training ground truth classes to be known. Thus, they are unapplicable in our setting.

\section{Experiments}
\label{sec:experiments}

Ensembling is used to improve accuracy and calibration. Calibrated predictions provide an invaluable tool for safe deployment, when a data distribution may deviate from the training distribution. Therefore, we evaluate the proposed SDS on the distributional shift and out-of-distribution detection datasets~\citep{ovadia2019can}. Full implementation details are given in Appendix E.

We use MNIST~\citep{lecun1998gradient}, CIFAR10/100~\citep{krizhevsky2009learning}, and ImageNet-1K~\citep{russakovsky2015imagenet} datasets with a distributional shift by providing the corrupted versions of validation data. For MNIST we use the rotated validation data with the angle of rotation varying from \(0^{\circ}\) to \(180^{\circ}\). For CIFAR10/100 and ImageNet we use corruptions from \citet{hendrycks2019robustness}. Results on validation data without corruption can be found in Appendix G. Additionally, we evaluate SDS on out-of-distribution (OOD) detection. We consider the pairs of in/out-distributions as: CIFAR10/100, CIFAR10/SVHN~\citep{netzer2011reading}, CIFAR100/10, CIFAR100/SVHN, where we vary the ratio of OOD examples.

%For MNIST we train a neural network with two convolutional layers and two fully connected layers.  For CIFAR100 we use WideResNet28x10~\citep{zagoruyko2016wide} and for ImageNet we use ResNet50~\citep{he2016deep} networks similar to~\citet{ashukha2020pitfalls}, and we use provided pretrained weights from there.  For CIFAR10 we use the torchvision implementation of ResNet18~\citep{he2016deep}.
For all datasets we use an ensemble of 3 members and we repeat experiments 3 times% on each dataset using different pretrained weights
, i.e., taking \(3 * 3 = 9\) independently trained networks and combining them into ensembles. Different ensemble sizes are considered in Appendix H.

\begin{figure*}[!t]
    \centering
    \begin{subfigure}[t]{0.49\textwidth}
        \centering
        \includegraphics[scale=0.8]{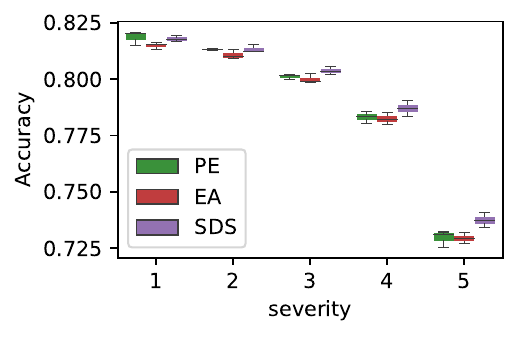}
        \caption{accuracy (\(\uparrow\))}
        \label{experiments:corrupted_cifar100_acc}
    \end{subfigure}
    \begin{subfigure}[t]{0.49\textwidth}
        \centering
        \includegraphics[scale=0.8]{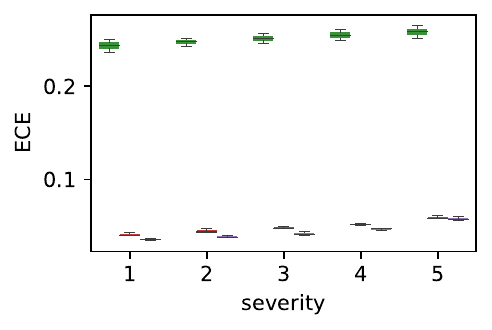}
        \caption{ECE (\(\downarrow\))}
        \label{experiments:corrupted_cifar100_ece}
    \end{subfigure}
    \hfill
    \begin{subfigure}[t]{0.49\textwidth}
        \centering
        \includegraphics[scale=0.8]{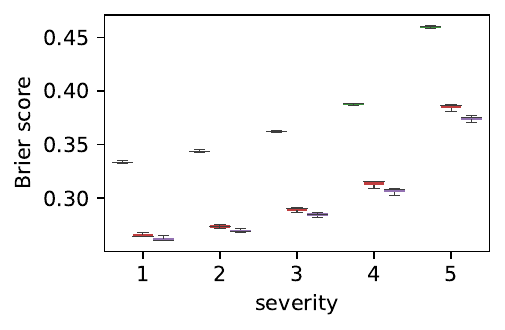}
        \caption{Brier Score (\(\downarrow\))}
        \label{experiments:corrupted_cifar100_brier}
    \end{subfigure}
    \begin{subfigure}[t]{0.49\textwidth}
        \centering
        \includegraphics[scale=0.8]{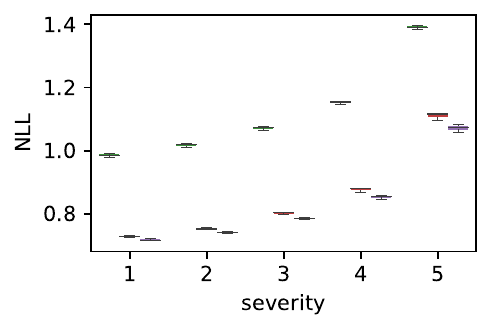}
        \caption{NLL (\(\downarrow\))}
        \label{experiments:corrupted_cifar100_nll}
    \end{subfigure}
    \caption{Results on CIFAR100 with Brightness corruption of increasing severity: (\subref{experiments:corrupted_cifar100_acc}) accuracy, (\subref{experiments:corrupted_cifar100_ece}) ECE, (\subref{experiments:corrupted_cifar100_brier}) Brier score, (\subref{experiments:corrupted_cifar100_nll}) NLL.}
    \label{experiments:corrupted_cifar100}
    %TODO bright results are better. Figure out how to make them visible
\end{figure*}

We compare 5 methods: DS~\citep{dawid1979maximum}, BCC~\citep{simpson2013dynamic}, EA~\citep{lakshminarayanan2016simple}, probabilistic ensembling (PE)~\citep{wang2023diversity} and SDS. DS and BCC use the hard labels from ensemble members, whereas EA and SDS use the soft labels. The full approach from~\citet{wang2023diversity} would be unfair as it includes training different ensemble members. We employ the plug-and-play version of PE, where only MoG is built on Laplace approximations of the existing ensemble members. It is worth noting that PE still has unfair advantage as Laplace approximation requires access to training labels, while all the rest of methods do not use any labels and training data in general. Also, PE is much more computationally expensive in comparison to other methods. Full computational costs are provided in Appendix I. Moreover, we cannot apply PE to ImageNet data as it requires 977 GiB of GPU memory.

We also consider an online version of SDS where we fix confusion matrices after estimating them (without supervision) on a batch of data and then we infer only true labels via E step for the rest of the data which can be done in a streaming way (Appendix F).

We evaluate accuracy, expected calibration error (ECE)~\citep{naeini2015obtaining}, Brier score~\citep{brier1950verification}, and negative log-likelihood (NLL). The 3 latter metrics are used for measuring calibration. For the OOD experiment we measure the area under receiver operating characteristic curve (AUROC).

 %For use in real life during inference we would use the already fitted parameters of the aggregating algorithms to predict the probabilities of the test data, thus computational overhead is low, although this case is not covered by experiments.

\subsection{Rotated MNIST}
The results for rotated MNIST are present in \figurename~\ref{experiments:mnist}.  In terms of accuracy all methods show similar performance. Hard label DS and BCC have also similar ECE and Brier score, that are worse than the results of both soft label EA and SDS. BCC, however, outperforms EA in terms of NLL for large angles of rotations. SDS outperforms DS, BCC, and EA in all 3 calibration metrics. This confirms that using both soft labels and confusion matrix estimation is beneficial for calibrated prediction ensembling. PE shows the best results in calibration metrics, but it does that with much higher computational cost and use of training labels.

\subsection{CIFAR10}

% \begin{wrapfigure}{r}{0.5\textwidth}
    % \vspace{-40px}
    % \centering
    \begin{figure}[tb]
        \includegraphics[scale=0.8]{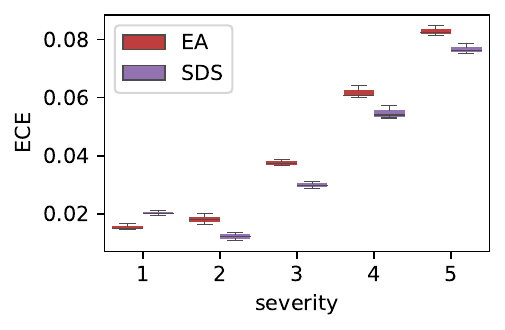}
    \caption{ECE (\(\downarrow\)) results on ImageNet with Zoom Blur corruption of increasing severity. The other 3 metrics have similar results for both methods.}
    \label{experiments:imagenet}

    % TODO zoom is better but need to make it visible
    \end{figure}
% \end{wrapfigure}

\begin{figure*}[!t]
    \centering
    \begin{subfigure}[t]{0.49\textwidth}
        \centering
        \includegraphics[scale=0.8]{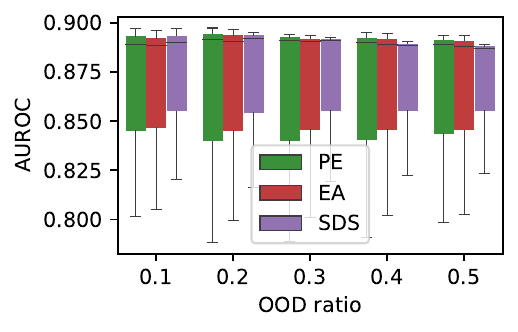}
        \caption{CIFAR10/100}
        \label{experiments:ood_cifar10_100}
    \end{subfigure}
    \begin{subfigure}[t]{0.49\textwidth}
        \centering
        \includegraphics[scale=0.8]{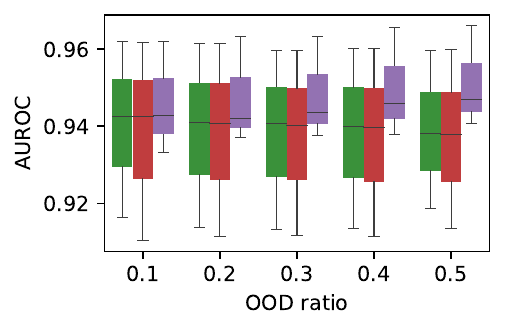}
        \caption{CIFAR10/SVHN}
        \label{experiments:ood_cifar10_svhn}
    \end{subfigure}
    \hfill
    \begin{subfigure}[t]{0.49\textwidth}
        \centering
        \includegraphics[scale=0.8]{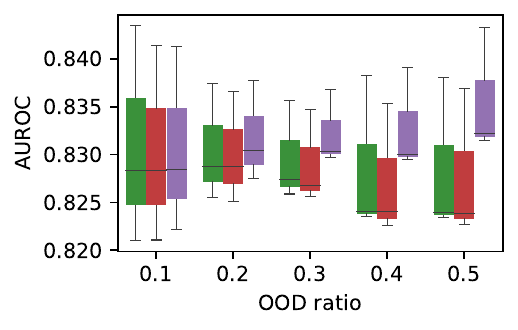}
        \caption{CIFAR100/10}
        \label{experiments:ood_cifar100_10}
    \end{subfigure}
    \begin{subfigure}[t]{0.49\textwidth}
        \centering
        \includegraphics[scale=0.8]{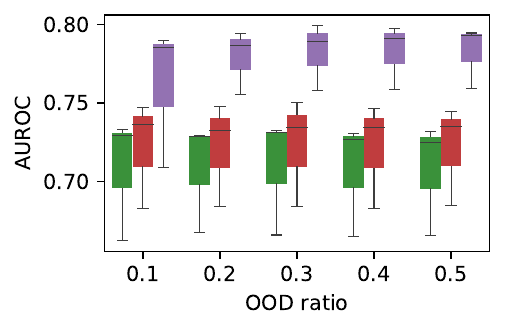}
        \caption{CIFAR100/SVHN}
        \label{experiments:ood_cifar100_svhn}
    \end{subfigure}
    \caption{AUROC (\(\uparrow\)) results for OOD of increasing ratio of OOD examples in the validation dataset.
    % : (\subref{experiments:ood_cifar10_100}) CIFAR10/100, (\subref{experiments:ood_cifar10_svhn}) CIFAR10/SVHN, (\subref{experiments:ood_cifar100_10}) CIFAR100/10, (\subref{experiments:ood_cifar100_svhn}) CIFAR100/SVHN pairs.
    }
    \label{experiments:ood}
\end{figure*}

The results for ``Frosted Glass Blur'' corruption on CIFAR10 are in \figurename~\ref{experiments:corrupted_cifar10}. Full results for all corruption types can be found in Appendix J.1. SDS has better results in terms of all 4 metrics. For all the other corruption types SDS has either similar or better results than EA and PE in all 4 metrics. DS and BCC demonstrate poorer performance than the other methods in the majority of cases.

\subsection{CIFAR100}

\figurename~\ref{experiments:corrupted_cifar100} shows the results for ``Brightness'' corruption on CIFAR100. All corruption type results can be found in Appendix J.2 (for visibility we exclude poor calibration DS and BCC results from the figure but they can be found in Appendix). SDS provides the best results in all calibration metrics and for most of severity levels in accuracy. For the other corruption types in comparison to EA, SDS shows similar results in accuracy and NLL, while SDS has better performance in Brier score and ECE than EA. In \(19\%\) cases EA outperforms in Brier and ECE for some severity levels. However, the absolute values for both methods in these cases are very small, therefore, SDS still provides good results. PE outperforms SDS in only \(35\%\) cases in calibration metrics and it achieves this at a much higher computational cost and requires the use of training labels.

% \begin{table*}[!th]
%     \centering
%     \begin{tabular}{lllll}
%         \toprule
%         Method & Accuracy \(\uparrow\) & ECE \(\downarrow\)            & Brier Score \(\downarrow\)     & NLL \(\downarrow\)    \\
%         \midrule
%         EA     & \(0.1919 \pm 0.0044\) & \(9.23e^{-2} \pm 0.72e^{-2}\) & \(-7.22e^{-2} \pm 0.63e^{-2}\) & \(4.6161 \pm 0.0707\) \\
%         SDS    & \(0.1919 \pm 0.0044\) & \(8.67e^{-2} \pm 0.72e^{-2}\) & \(-7.38e^{-2} \pm 0.61e^{-2}\) & \(4.6077 \pm 0.0667\) \\
%         \bottomrule
%     \end{tabular}
%     \caption{Results on corrupted ImageNet with ``Frosted Glass Blur'' corruption of severity 3.}
%     \label{tab:corrupted_imagenet}
% \end{table*}

\subsection{ImageNet}

\figurename~\ref{experiments:imagenet} presents the ECE results on ImageNet with ``Zoom Blur'' corruption. We present results of EA and SDS for ECE only as for the other metrics performances of both methods are very similar (we excluded DS and BCC for better visibility due to poor performance of these methods). Full results including DS and BCC for all corruption types are in Appendix J.3. For the majority of corruption types SDS outperforms EA in ECE. All other metrics have similar results for both methods.

%We also investigate SDS performance with respect to different ensemble sizes on ImageNet. The results can be found in Appendix. We show that the comparison of SDS and EA is consistent across different number of ensemble members.

%Results for ``Frosted Glass Blur''~\citep{hendrycks2019robustness} corrupted ImageNet with severity 3 are given in Table~\ref{tab:corrupted_imagenet}. Here ensembles of size 5 are used. We can observe that while both EA and SDS have the same accuracy, SDS outperforms EA in all 3 calibration metrics.

\subsection{Out-of-distribution Detection}
 The OOD results are in~\figurename~\ref{experiments:ood} (due to previous poor results we exclude DS and BCC). PE and EA show similar results in 3 dataset pairs while EA shows better performance on CIFAR100/SVHN pair. SDS shows similar AUROC on CIFAR10/100 pair and outperforms both PE and EA on other dataset pairs especially as the ratio of OOD examples increases. This shows that uncertainties estimation provided by SDS is more useful for this downstream task of OOD detection.

\section{Conclusions}
\label{sec:conclusions}
We present the Soft Dawid Skene method for ensembling (and crowdsourcing). Compared to EA, it additionally estimates the confusion matrices thus improving aggregated predictions by weighing ensemble members according to their inferred performance. Compared to DS, it operates with soft labels, which improves calibration of predictions by incorporating uncertainty of individual ensemble members. Inclusion of soft labels in Dawid Skene model requires different priors and different inference methods.

We perform extensive evaluation of SDS on classification datasets with a distributional shift and OOD detection and show that SDS demonstrates similar or better performance than the state-of-the-art EA both in terms of accuracy and calibration of the predictions. As we discussed above, EA is the only existing alternative for aggregation of ensemble members' predictions that does not use training data. Therefore, for the first time we provide another competitive option for ensemble aggregation.

We only provide the empirical results for ensembles where each ensemble member is a neural network of the same architecture but trained independently on the same data. This is one of the most challenging settings for the model as it violates the assumption of conditional independence of ensemble members. We expect SDS to demonstrate even larger difference with EA for the settings where ensemble members are less correlated (for example, when coupled with bagging or knowledge distillation). We will explore these cases in the future.

\appendix
\section{Introduction}

This appendix provides technical details about the experiments and additional results for the main paper.

\section{Notation}
Here we recall the main notation from the paper:
\begin{itemize}
    \item \(N\) --- the number of data points in a dataset;
    \item \(K\) --- the number of classifiers, i.e., ensemble members;
    \item \(J\) --- the number of classes;
    \item \(\mathbf{t} = \{t_i\}_{i=1}^N, t_i \in \{1, \ldots, J\}\) --- true labels of the data points (hidden);
    \item \(\mathbf{C} = \{\mathbf{c}_i\}_{i=1}^N, \mathbf{c_i} = \{\mathbf{c}_i^{(k)}\}_{k=1}^K, \mathbf{c}_i^{(k)} = \{c_{il}^{(k)}\}_{l=1}^J \in [0, 1]^J\) --- predicted probabilities of class labels given by the ensemble members;
    \item \(\boldsymbol{\Pi} = \{\boldsymbol{\Pi}^{(k)}\}_{k= 1}^K, \boldsymbol{\Pi}^{(k)} = \{\pi_{jl}^{(k)}\}_{j, l=1}^J \in [0, \infty)^{J \times J}\) --- confusion matrices (more precisely, these are parameters of the prior Dirichlet distribution of the actual confusion matrices, but we refer to \(\boldsymbol{\Pi}^{(k)}\) as a confusion matrix for simplicity);
    \item \(\boldsymbol{\nu} = \{\nu_j\}_{j=1}^J \in [0, 1]^J\) --- prior multinomial distribution of the true labels.
\end{itemize}

\section{Illustrative example of how SDS works in practice}

In this section we provide an example how SDS works in practice and how the use of confusion matrices and soft labels allows it to outperform baselines.

Let us consider a sample (one data point) from CIFAR100 with Brightness corruption at severity level 3 (section 6.3 of the main paper). For this sample SDS predicts the correct class label and EA does not. Hereafter, we refer to the softmax output as the model's confidence.

This data sample (number 35 in the validation dataset) belongs to class 73. The three ensemble members predict classes 57, 95, and 38, respectively. EA predicts class 57 because ensemble member 1 is overconfident in its prediction (with a confidence of 0.22), which dominates the votes from the other two members (whose maximum confidences are 0.12 and 0.13, respectively).

SDS, however, predicts the correct class, 73. Its decision is more nuanced than that of EA. The prediction of a class label is the maximum of probabilities given in eq. (3) in the main paper:
\[
\log p(t_i=j | \mathbf{C}, \boldsymbol\nu, \boldsymbol\Pi) \propto  \log\nu_j + \sum_{k, l} (\pi_{j l}^{(k)}-1) \log c_{il}^{(k)} - \sum_k \left( \sum_l \log \Gamma(\pi_{j l}^{(k)}) - \log \Gamma(\sum_l \pi_{j l}^{(k)})\right).
\]
Let us focus on the second term of this equation (which determines the class prediction in this case):

\begin{equation}
\label{eq:second_term}
\sum_{k, l} (\pi_{jl}^{(k)} - 1)\log c_{il}^{(k)}. \tag{A.1}
\end{equation}
\newpage
The other two terms of eq. (3) are similar for different $j$ and thus do not define the class label.

First, we need to notice that all soft labels (class probabilities) from each ensemble member are used in the decision, not just the class with the maximum confidence, as it is usually the case for EA.

Now, let us interpret the confusion matrices learned by SDS. When class 73 is the correct class, ensemble member 1, according to confusion matrices learnt by SDS, would output classes [23, 30, \textbf{67}, 72, \textbf{73}, 87, 91, \textbf{95}] with higher probabilities (higher than average).

If we check the top-10 classes predicted by ensemble member 1 for our sample, they are [57, 69, 65, 38, \textbf{67}, 27, \textbf{95}, 79, 32, \textbf{73}]. The highlighted class labels appear in both lists, meaning they will significantly contribute in eq.~(\ref{eq:second_term}) as votes for class 73.

Similarly, when class 73 is the correct class, ensemble member 2, according to SDS, would output classes [\textbf{30}, 67, \textbf{73}, 93, \textbf{95}] with higher probabilities.

For our data sample, ensemble member 2 outputs the top-10 classes as: [\textbf{95}, \textbf{73}, 74, 44, \textbf{30}, 27, 72, 38, 50, 19]. Thus, three elements in the sum from eq.~(\ref{eq:second_term}) significantly contribute to the vote for class 73.

Finally, when class 73 is the correct class, SDS has learned that ensemble member 3 would output classes [10, 23, 30, 44, 62, 67, 70, \textbf{73}, 84, 91, 93, \textbf{95}].

The top-10 classes that ensemble member 3 predicts for the given data sample are: [38, 72, \textbf{95}, \textbf{73}, 69, 57, 34, 74]. Two elements would significantly contribute to the vote for class 73 in eq.~(\ref{eq:second_term}).

In fact, these contributions are enough for SDS to predict the correct class, even though none of the ensemble members individually guesses the correct class.

From this example, we can see that SDS improves the performance by looking at all soft outputs from ensemble members combined with confusion matrices, i.e. learnt mistake patterns, for these ensemble members.

\section{Hyperparameters}
\label{sec:hyperparameters}
Our proposed EM algorithm for Soft Dawid Skene (SDS) model has the following hyperparameters:
\begin{itemize}
    \item \(\alpha\) --- the weight of the new estimate of \(\mathbf{t}\) in Polyak averaging in E-step of the algorithm (see eq.(11) in the main text);
    \item number of iterations of the full EM algorithm --- though in theory it can be run until the convergence of the \(Q\) function, we find in practice it is enough to run it for the limited number of iterations;
    \item parameters of the optimiser used for M-step for optimising \(\boldsymbol{\Pi}\):
    \begin{itemize}
    \item type of optimiser. Any optimiser is suitable here, we use AdamW~\citep{loshchilov2017decoupled} in all our experiments;
    \item learning rate of the optimiser. We found that \(10^{-4}\) provided good results for all datasets;
    \item weight decay. Our initial experiments showed that the actual value of this parameter does not matter much as long as the weight decay is present, so we set it to \(10^{-4}\) for all experiments;
    \item number of iterations of the optimiser --- since this optimisation is internal inside the outer EM loop, we do not need to run the optimiser untilconvergence at every M-step. Our initial experiments showed that a small number of iterations (i.e. \(5\)) already provided sufficient results. We use \(5\) for MNIST, CIFAR10/100. For ImageNet experiments even \(1\) iteration of the optimiser was sufficient.
    \end{itemize}
\end{itemize}

For \(\alpha\) we use the range \([10^{0}, 10^{-1}, 10^{-2}, 10^{-3}, 10^{-4}, 10^{-5}, 10^{-6}]\) for tuning. We also incorporated simple scheduling for \(\alpha\), where we changed its value once during the training. We optimise both when the change happens and to which value \(\alpha\) changes.

It is worth noting that \(\alpha = 1\) corresponds to the case when no Polyak averaging is used during the E step of the algorithm. By using it during tuning we simultaneously conducted an ablation study on whether Polyak averaging is required. For all datasets we have observed a very poor performance with \(\alpha = 1\), therefore, we have confirmed that this step is essential for our algorithm.

For the number of iterations we set initially the number of iterations to be large (e.g., 500), and then choose the final number based on the historical performance over the number of iterations. Since we use 4 different metrics for comparison, which are not necessarily optimised with the same values of hyperparameters, we manually assess the result of different hyperparameter values and choose values that provide a good trade-off for all 4 metrics.

The final hyperparameter values we use in the experiments are given in Table~\ref{tab:hyperparameters}. Here, the list of values for \(\alpha\) means that we apply the change of values for \(\alpha\) during the EM iterations and the second row of the table specifies at which iteration this change occurs.

\begin{table*}[!th]
    \centering
    \caption{Hyperparameter values used in the experiments}
    \label{tab:hyperparameters}
    \begin{tabular}{lllll}
        \toprule
        Hyperparameter & MNIST & CIFAR-10            & CIFAR-100     & ImageNet   \\
        \midrule
        \(\alpha\)     & \(10^{-3}\) & \(10^{-3}\) & \([10^{-5}, 10^{-3}]\) & \([10^{-3}, 10^{-5}]\) \\
        iteration number for \(\alpha\) change & --- & --- & 100 & 20 \\
        number of EM iterations    & \(100\) & \(100\) & \(200\) & \(25\) \\
        \bottomrule
    \end{tabular}
\end{table*}

\section{Experimental Details}
\label{sec:experimental_details}

All experiments are conducted on 24GB NVIDIA Titan RTX GPU / Intel Core i9 10 Core CPU / 128 GB RAM machine with the Ubuntu operating system. We use PyTorch~\citep{paszke2019pytorch} for implementation.
% TODO add details on OOD
\subsection{Datasets} We evaluate performance of SDS on different vision classification datasets.
\begin{itemize}
    \item MNIST~\citep{lecun1998gradient}: a dataset of handwritten digit recognition with greyscale images of size \(28 \times 28\) pixels. Each image contains a digit that should be classified into one of \(10\) classes. For a corrupted version of validation data we rotate the images with angles of rotation ranging from \(0^{\circ}\) to \(180^{\circ}\);
    \item CIFAR10/100~\citep{krizhevsky2009learning}: a dataset of \(32 \times 32\) colour images with 10/100 classes respectively. For corrupted versions of validation data we use corruptions from~\citet{hendrycks2019robustness}. These are 15 different corruptions ranging from Gaussian noise to fog weather, and each corruption type has 5 severity levels;
    \item ImageNet-1K~\citep{russakovsky2015imagenet}: a dataset of colour images with 1000 classes. For corrupted version of validation data we use corruptions from~\citet{hendrycks2019robustness}.
\end{itemize}

Ensemble members are trained on the clean training data and then evaluated on the corrupted validation data.

In our experiments we first train an ensemble of neural networks on training data independently of each other with different random seeds, then we fit the aggregation algorithms on validation data and evaluate metrics on validation data.  Aggregating algorithms do not use true labels to fit, therefore there is no overfitting in this procedure. PE as exception also includes the step of fitting Laplace approximation of each ensemble member on training data.

We further evaluate performance of SDS on OOD tasks. For this we use an ensemble with members trained on \(X\) and validation data from \(X\) as in-distribution examples and \(Y\)  as out-of-distribution examples. For this experiment uncorrupted versions of datasets are employed. We take the validation data from \(X\) and at random replace a specified ratio of data points with random images from \(Y\). We vary the ratio of these OOD examples as \([0.1, 0.2, 0.3, 0.4, 0.5]\). The pairs of \(X/Y\) are considered: CIFAR10/100, CIFAR10/SVHN~\citep{netzer2011reading}, CIFAR100/10, CIFAR100/SVHN. SVHN: a dataset of \(32 \times 32\) colour images with 10 classes.

\subsection{Neural Network Ensemble Members}
\label{sec:exp_ens_members}
\subsubsection{MNIST}
For MNIST we train 9 neural networks with the following architecture:
\begin{verbatim}
torch.nn.Sequential(
    torch.nn.Conv2d(1, 32, 3),
    torch.nn.ReLU(),
    torch.nn.Conv2d(32, 64, 3),
    torch.nn.ReLU(),
    torch.nn.MaxPool2d(2),
    torch.nn.Dropout(0.25),
    torch.nn.Flatten(),
    torch.nn.Linear(9216, 128),
    torch.nn.ReLU(),
    torch.nn.Dropout(0.25),
    torch.nn.Linear(128, 10),
)
\end{verbatim}
We train each network for 10 epochs, batch size 24, AdamW optimiser~\citep{loshchilov2017decoupled}, learning rate \(10^{-3}\). No transformation is applied for the data except normalisation. We then combine these 9 neural networks into 3 ensembles of size 3.

\subsubsection{CIFAR10}
For CIFAR10 we train 9 Resnet18~\citep{he2016deep} neural networks (torchvision implementation). We train each network for 40 epochs, batch size 128, stochastic gradient descent (SGD) optimiser, with \(10^{-1}\) initial learning rate, momentum 0.9, weight decay \(10^{-4}\). Learning rate has a multistep PyTorch scheduler with decay at epochs 20 and 30 by \(\gamma=0.1\). During training random rotations (with degree \(90^{\circ}\)), random horizontal flip, and normalisation transformations are applied. Only normalisation is applied during inference. We then combine these 9 neural networks into 3 ensembles of size 3.

\subsubsection{CIFAR100}
For CIFAR100 we use 9 WideResNet28x10~\citep{zagoruyko2016wide} networks, which trained weights we take from~\citet{ashukha2020pitfalls}. Only normalisation transformation is applied during inference. We then combine these 9 neural networks into 3 ensembles of size 3.

\subsubsection{ImageNet}
For ImageNet we use 9 ResNet50~\citep{he2016deep} networks with pretrained weights from~\citet{ashukha2020pitfalls}. Centre crop with size 224 and normalisation transforms are applied during inference. We then combine these 9 neural networks into 3 ensembles of size 3.

\subsection{Ensembling method implementations}

For DS we use the implementation from~\citet{ustalov2021learning}\footnote{\url{https://github.com/Toloka/crowd-kit}} (license Apache). PE implementation uses Laplace approximation from the \textit{laplace} library~\citep{daxberger2021laplace}\footnote{\url{https://github.com/AlexImmer/Laplace/}} (license MIT).

\subsection{Metrics} We use the following metrics for comparison:
\begin{itemize}
    \item Accuracy
    \item Expected calibrated error (ECE)~\citep{naeini2015obtaining}, which is a weighted difference of prediction accuracy and confidence of predictions.%:
    % \begin{equation}
    %     ECE = \sum_{s=1}^S \frac{|B_s|}{N}|\text{acc}(B_s) - \text{conf}(B_s)|,
    % \end{equation}
    % where \(B_s\) is \(s\)-th bin of \(S\) bins of predicted probabilities, \(\text{acc}(B_s) = \frac{1}{|B_s|}\sum_{i\in B_s} \mathbb{I}[t_i = \hat{t}_i]\), and \(\text{conf}(B_s) = \frac{1}{|B_s|}\sum_{i\in B_s}p(\hat{t}_i)\) with \(\mathbb{I}\) being the indicator function, which is 1 when the statement within the brackets is true and 0 otherwise, \(\hat{t}_i = \text{argmax}\, p(t_i)\), \(t_i\) denotes the ground truth label and \(p(t_i)\) is a probability distribution predicted by a model.
    We use 300 bins for ECE computation in all experiments.
    \item Brier score~\citep{brier1950verification}, which is the mean difference between the predicted probabilities and the actual outcome
    \item Negative log likelihood (NLL)
\end{itemize}

For evaluation metrics we adapt the implementation from~\citet{ovadia2019can}\footnote{\url{https://github.com/google-research/google-research/blob/master/uq_benchmark_2019/metrics_lib.py}} (license Apache).

\section{Additional results on online SDS}
In this section, we explore an online version of SDS. In the vanilla SDS, as presented earlier, a batch of data is required for the model to learn confusion matrices and class probabilities. Here, we compare it with an online variant. First, we run the vanilla SDS on a subset of test data to learn prior class probabilities, $\boldsymbol{\nu}$ and confusion matrices, $\boldsymbol{\Pi}$ (a learning step). We then fix these learned parameters. For the rest of test data, we only run one E-step of SDS to infer predictions (an inference step). This allows the inference step to be applied to streaming data, processing one data point at a time. Importantly, the entire online version, including both learning and inference, remains fully unsupervised.

\begin{figure}[ht]
\centering
        \includegraphics[scale=0.9]{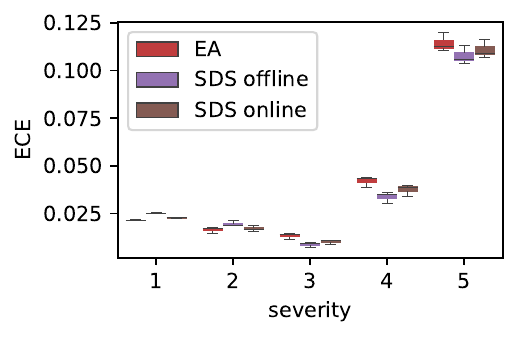}
    \caption{ECE (\(\downarrow\)) results for online SDS on ImageNet with Fog corruption of increasing severity.}
    \label{experiments:online}
    % TODO zoom is better but need to make it visible
\end{figure}

\figurename~\ref{experiments:online} presents comparison of the vanilla SDS (offline), online SDS and EA on ImageNet with ``Fog'' corruption in terms of ECE. We first randomly select 10,000 data points out of validation dataset and run the vanilla SDS on this subset of data. We then fix prior class probabilities $\boldsymbol{\nu}$ and confusion matrices $\boldsymbol{\Pi}$ and apply only E-step on the remaining 40,000 validation data points. It makes the online version significantly faster. It takes around 22 seconds for the learning step (on 10,000 data points) and 1 second for the inference step (on 40,000 data points). By comparison, vanilla SDS takes 82 seconds for 40,000 data points and 114 seconds for the full 50,000 points. (A full computational cost analysis of SDS and baselines is provided in Appendix~\ref{sec:cost}.)

We also apply EA and another instance of vanilla SDS (offline SDS) to the 40,000 data points. The ECE results are given in~\figurename~\ref{experiments:online}. We can see that the online SDS provides results in between of EA and offline SDS. It still gives the performance gain for higher levels of severities, but does it on a fraction of cost of offline SDS. Overall, the online version offers a trade-off between calibration performance and computational efficiency.

\section{Additional results on within datasets}
\label{sec:within_results}
Here we provide the results for original validation data without corruptions. For MNIST it is included in~\figurename~2 of the main paper as the angle equal \(0\).

\begin{figure*}[!th]
    \centering
    \begin{subfigure}[t]{0.49\textwidth}
        \centering
        \includegraphics[scale=0.75]{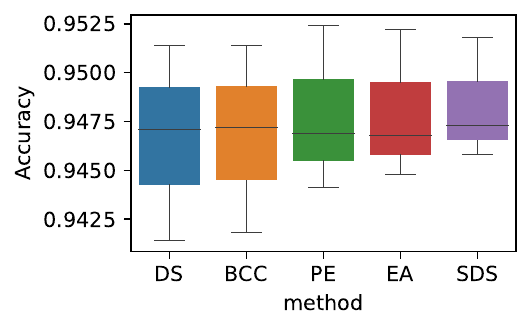}
        \caption{accuracy (\(\uparrow\))}
        \label{experiments:clean_cifar10_acc}
    \end{subfigure}
    \begin{subfigure}[t]{0.49\textwidth}
        \centering
        \includegraphics[scale=0.75]{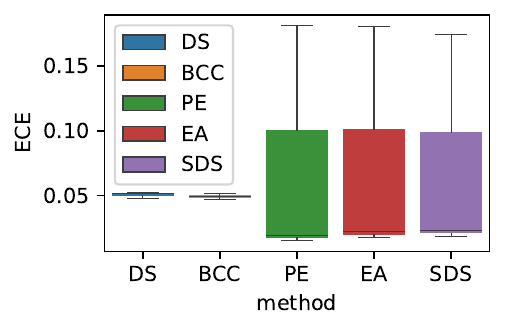}
        \caption{ECE (\(\downarrow\))}
        \label{experiments:clean_cifar10_ece}
    \end{subfigure}
    \hfill
    \begin{subfigure}[t]{0.49\textwidth}
        \centering
        \includegraphics[scale=0.75]{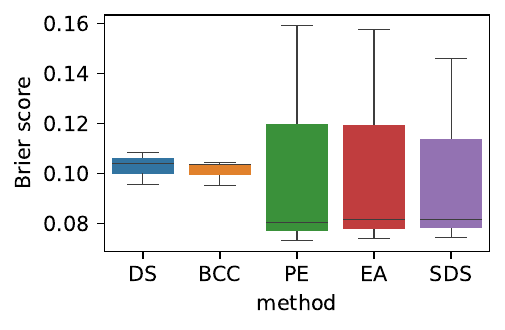}
        \caption{Brier Score (\(\downarrow\))}
        \label{experiments:clean_cifar10_brier}
    \end{subfigure}
    \begin{subfigure}[t]{0.49\textwidth}
        \centering
        \includegraphics[scale=0.75]{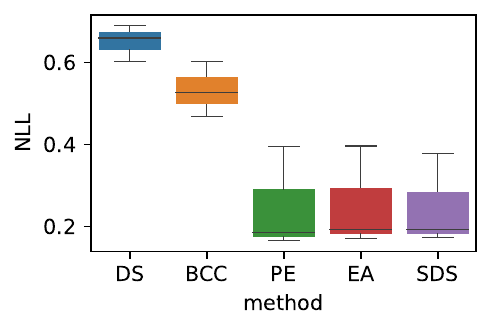}
        \caption{NLL (\(\downarrow\))}
        \label{experiments:clean_cifar10_nll}
    \end{subfigure}
    \caption{Results on CIFAR10 without corruptions.}
    \label{experiments:clean_cifar10}
\end{figure*}

\begin{figure*}[!th]
    \centering
    \begin{subfigure}[t]{0.49\textwidth}
        \centering
        \includegraphics[scale=0.75]{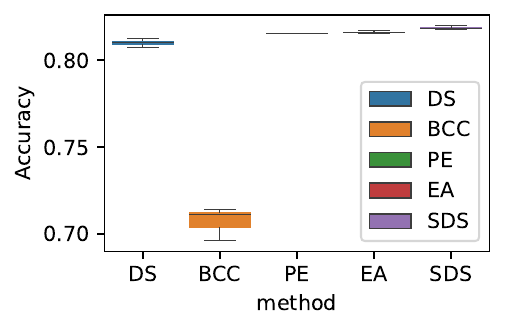}
        \caption{accuracy (\(\uparrow\))}
        \label{experiments:clean_cifar100_acc}
    \end{subfigure}
    \begin{subfigure}[t]{0.49\textwidth}
        \centering
        \includegraphics[scale=0.75]{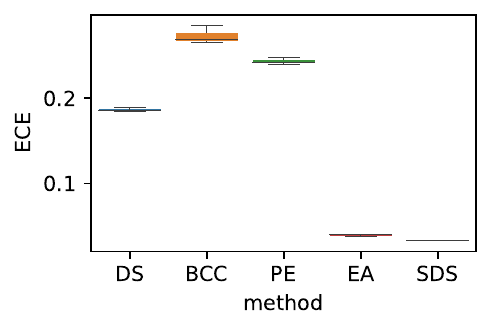}
        \caption{ECE (\(\downarrow\))}
        \label{experiments:clean_cifar100_ece}
    \end{subfigure}
    \hfill
    \begin{subfigure}[t]{0.49\textwidth}
        \centering
        \includegraphics[scale=0.75]{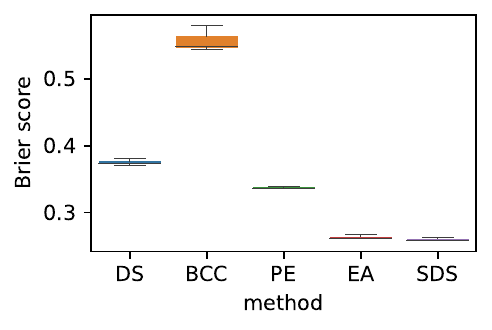}
        \caption{Brier Score (\(\downarrow\))}
        \label{experiments:clean_cifar100_brier}
    \end{subfigure}
    \begin{subfigure}[t]{0.49\textwidth}
        \centering
        \includegraphics[scale=0.75]{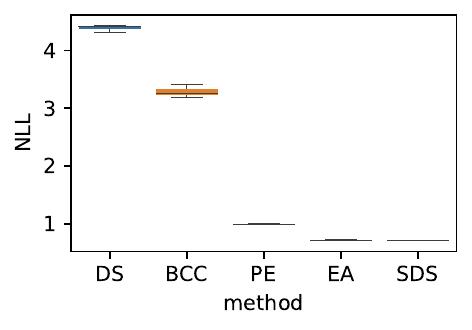}
        \caption{NLL (\(\downarrow\))}
        \label{experiments:clean_cifar100_nll}
    \end{subfigure}
    \caption{Results on CIFAR100 without corruptions.}
    \label{experiments:clean_cifar100}
\end{figure*}

\begin{figure*}[!th]
    \centering
    \begin{subfigure}[t]{0.49\textwidth}
        \centering
        \includegraphics[scale=0.75]{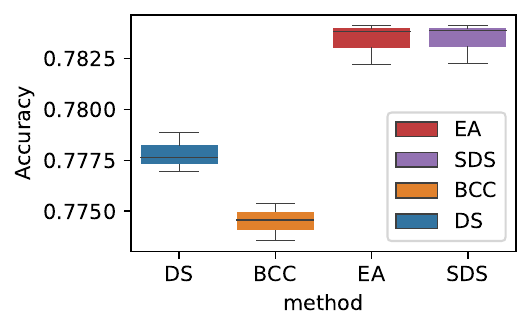}
        \caption{accuracy (\(\uparrow\))}
        \label{experiments:clean_imagenet_acc}
    \end{subfigure}
    \begin{subfigure}[t]{0.49\textwidth}
        \centering
        \includegraphics[scale=0.75]{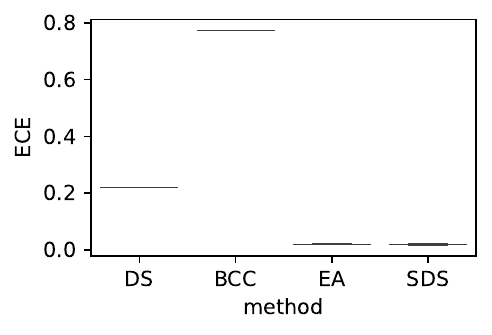}
        \caption{ECE (\(\downarrow\))}
        \label{experiments:clean_imagenet_ece}
    \end{subfigure}
    \hfill
    \begin{subfigure}[t]{0.49\textwidth}
        \centering
        \includegraphics[scale=0.75]{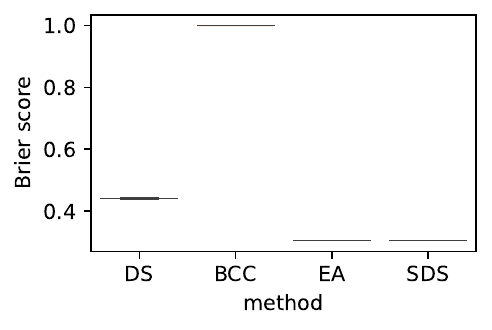}
        \caption{Brier Score (\(\downarrow\))}
        \label{experiments:clean_imagenet_brier}
    \end{subfigure}
    \begin{subfigure}[t]{0.49\textwidth}
        \centering
        \includegraphics[scale=0.75]{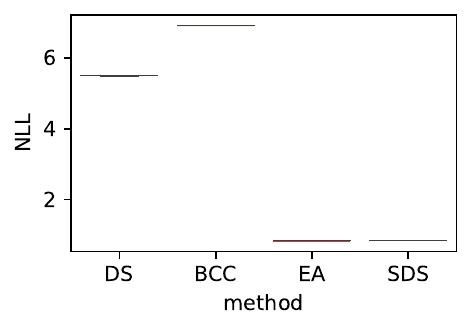}
        \caption{NLL (\(\downarrow\))}
        \label{experiments:clean_imagenet_nll}
    \end{subfigure}
    \caption{Results on ImageNet without corruptions.}
    \label{experiments:clean_imagenet}
\end{figure*}

Figures~\ref{experiments:clean_cifar10}--\ref{experiments:clean_imagenet} present these results for CIFAR10, CIFAR100 and ImageNet, respectively. On CIFAR10 all methods show similar performance with hard-label DS and BCC having less variance between independent runs and underperforming in terms of NLL. The results on CIFAR100 demonstrate advantages of EA and SDS in terms of all 3 calibration metrics, and ImageNet highlights superiority of EA and SDS in all 4 metrics measured.

\section{Additional results on varying ensemble sizes}
\label{sec:varying_ens_size}

Here, we provide the results for comparing different ensemble sizes. For illustrative purposes we show the results of EA (the main baseline) and SDS on CIFAR-100 data with Gaussian Noise corruption with severity 3 (Table~\ref{tab:cifar100_vary_ens_size}) and Imagenet with Frost corruption with severity 3 (Table~\ref{tab:imagenet_vary_ens_size}). The trend persists for other data. In general, performance slowly increases with the size of the ensembles. Relative performance between different methods stays mostly the same, some differences become significant with the increase of the ensemble size.

\begin{table*}[!th]
    \centering
    \caption{Results on CIFAR100 with varying ensemble size for Gaussian Noise corruption of severity~3. The mean and std are computed for 3 independent Monte Carlo runs.}
    \label{tab:cifar100_vary_ens_size}
    \begin{tabular}{lllllllll}
        \toprule
        Metric & Method & Ensemble size 3 & Ensemble size 5 & Ensemble size 7 & Ensemble size 10\\
        \midrule
        Accuracy & EA & 0.116 $\pm$ 0.015 & 0.121 $\pm$ 0.010 & 0.123 $\pm$ 0.003 & 0.124 $\pm$ 0.004\\
          & SDS & 0.116 $\pm$ 0.015 & 0.121 $\pm$ 0.010 & 0.123 $\pm$ 0.003 & 0.124 $\pm$ 0.004 \\
        ECE & EA & 0.476 $\pm$ 0.035 & 0.410 $\pm$ 0.013 & 0.386 $\pm$ 0.017  & 0.396 $\pm$ 0.017\\
          & SDS & \textbf{0.419} $\pm$ 0.033 & \textbf{0.359} $\pm$ \textbf{0.012} & \textbf{0.337} $\pm$ \textbf{0.016} & \textbf{0.346} $\pm$ \textbf{0.016} \\
        Brier score & EA & 1.280 $\pm$ 0.039 & 1.209 $\pm$ 0.017 & 1.182 $\pm$ 0.020 & 1.193 $\pm$ 0.017\\
          & SDS & \textbf{1.213} $\pm$ 0.033 & \textbf{1.157} $\pm$ \textbf{0.013} & \textbf{1.135} $\pm$ \textbf{0.017} & \textbf{1.144} $\pm$ \textbf{0.014} \\
        NLL & EA & 5.714 $\pm$ 0.265 & 5.374 $\pm$ 0.278 & \textbf{5.113} $\pm$ 0.254 & \textbf{5.137} $\pm$ 0.239\\
          & SDS & \textbf{5.611} $\pm$ 0.212 & \textbf{5.365} $\pm$ 0.240 & 5.115 $\pm$ 0.252 & 5.139 $\pm$ 0.246\\
	\bottomrule
    \end{tabular}
\end{table*}

\begin{table*}[!th]
    \centering
    \caption{Results on ImageNet with varying ensemble size for Frost corruption of severity~3. The mean and std are computed for 3 independent Monte Carlo runs.}
    \label{tab:imagenet_vary_ens_size}
    \begin{tabular}{llll}
        \toprule
        Metric & Method & Ensemble size 3 & Ensemble size 5 \\
        \midrule
        Accuracy & EA & 0.337 $\pm$ 0.001 & 0.345 $\pm$ 0.004 \\
          & SDS & 0.337 $\pm$ 0.001 & 0.345 $\pm$ 0.004 \\
        ECE & EA & 0.041 $\pm$ 0.003 & 0.022 $\pm$ 0.002 \\
          & SDS & \textbf{0.034} $\pm$ \textbf{0.003} & \textbf{0.018} $\pm$ \textbf{0.002} \\
        Brier score & EA & 0.794 $\pm$ 0.002 & 0.785 $\pm$ 0.003 \\
          & SDS & \textbf{0.793} $\pm$ 0.002 & 0.785 $\pm$ 0.003 \\
        NLL & EA & \textbf{3.478} $\pm$ \textbf{0.003} & \textbf{3.409} $\pm$ 0.025 \\
          & SDS & 3.492 $\pm$ 0.003 & 3.423 $\pm$ 0.025 \\
	\bottomrule
    \end{tabular}
\end{table*}

\section{Computational cost analysis}
\label{sec:cost}
In this section we provide the wall-clock computational cost comparison. We would like to emphasise that we have not made it our goal to optimise implementation of any of the methods we are comparing, therefore, the computational cost provided should not be taken as a true optimal cost.

DS, BCC, EA and SDS are all implemented in the same manner: ensemble members are run on validation data to make predictions (\textit{pred}), then the aggregation method is run on saved predictions (\textit{aggr}). The total run time (\textit{total}) for these 4 methods is then: \textit{pred} + \textit{aggr}, where \textit{pred} time would be exactly the same for all 4 methods as they use the same predictions.

PE is conceptually different and it is also implemented via the laplace library~\citep{daxberger2021laplace}, which makes it also different from the other methods. First, the Laplace approximation is run for each ensemble members on train data (\textit{lap}). Then predictive samples from these approximations are generated on validation data (\textit{pred}). The aggregation is done by taking the mean of predictions (\textit{aggr}). Therefore, the total run time (\textit{total}) for PE is: \textit{lap} + \textit{pred} + \textit{aggr}.

Tables~\ref{tab:comp_cost_1}--\ref{tab:comp_cost_2} present the wall-clock computational cost. The Laplace approximation step of PE makes it much slower than the other methods. Even on ImageNet data where we fail to run the prediction step due to the high memory demand because of the Laplace approximation PE would be the slowest method (only BCC has the higher \textit{total} time but any reasonable \textit{pred} time for PE would bring the \textit{total} cost higher).

EA being the simplest method is the fastest. SDS scales better for the large ImageNet dataset than both DS and BCC.

\begin{table*}[!th]
    \centering
    \caption{Computational cost comparison in seconds of different methods for ensemble of size 3 on MNIST and CIFAR10.}
    \label{tab:comp_cost_1}
\begin{tabular}{lcccc|cccc}
        \toprule
        \multirow{2}{*}{Method} & \multicolumn{4}{c}{MNIST} & \multicolumn{4}{c}{CIFAR10} \\
         & lap & pred & aggr & total & lap & pred & aggr & total\\
        \midrule
        DS & --- & 5.5 & 0.77 & 6 &  --- & 27 & 0.69 & 28 \\
        BCC & --- & 5.5 & 1.04 & 7 &  --- & 27 & 1 & 28 \\
        PE & 39 & 5 & 0.002 & 44 & 339 & 27 & 0.002 & 366 \\
        EA & --- & 5.5 & 0.001 & 6 & --- & 27 & 0.002 & 27 \\
        SDS & --- & 5.5 & 0.88 & 6 & --- & 27 & 0.89 & 28 \\
        \bottomrule
    \end{tabular}

\end{table*}

\begin{table*}[!th]
    \centering
    \caption{Computational cost comparison in seconds of different methods for ensemble of size 3 on CIFAR100 and ImageNet. X in PE results on ImageNet denote that we cannot run the prediction and aggregation steps.}
    \label{tab:comp_cost_2}
    \begin{tabular}{lcccc|cccc}
        \toprule
        \multirow{2}{*}{Method} & \multicolumn{4}{c}{CIFAR100}  & \multicolumn{4}{c}{ImageNet}  \\
         & lap & pred & aggr & total & lap & pred & aggr & total\\
        \midrule
        DS &  --- & 24 & 3 & 27 & --- & 285 & 165 & 450 \\
        BCC &  --- & 24 & 9 & 33 & --- & 285 & 321 & 606 \\
        PE & 339 & 70 & 0.01 & 409 & 597 & X & X & >597 \\
        EA &  --- & 24 & 0.05 & 24 & --- & 285 & 0.3 & 285 \\
        SDS & --- & 24 & 11 & 35 & --- & 285 & 114 & 399 \\
        \bottomrule
    \end{tabular}

\end{table*}

\section{Results for All Corruptions}

All tables provide the mean and std results for 3 independent Monte Carlo runs of each experiment.

\subsection{CIFAR10}
\label{sec:all_results_cifar10}
%TODO bold better results in tables

Results for all corruptions on CIFAR10 can be found in Tables~\ref{tab:cifar10_noise}--\ref{tab:cifar10_digital}. %For all corruptions and all 4 metrics SDS demonstrates either similar or better results than EA.
{
\scriptsize
\begin{longtable}{p{1cm}p{0.8cm}p{0.8cm}llllllll}
    \caption{Results on CIFAR10 with noise corruptions of increasing severity.}
    \label{tab:cifar10_noise}\\
        \toprule
        Corruption & Metric & Method & Severity 1 & Severity 2 & Severity 3 & Severity 4 & Severity 5 \\
        \midrule
        \endfirsthead
    \caption[]{Results on CIFAR10 with noise corruptions of increasing severity (continued).}
    \label{tab:cifar10_noise}\\
        \toprule
        Corruption & Metric & Method & Severity 1 & Severity 2 & Severity 3 & Severity 4 & Severity 5 \\
        \midrule
        \endhead
        \midrule
        \multicolumn{9}{r}{\textit{Continued on next page}}\\
        \endfoot
        \bottomrule
        \endlastfoot
        \multirow{20}{1cm}{Gaussian Noise} & \multirow{5}{0.8cm}{Accuracy} & DS & 0.676 $\pm$ 0.008 & 0.426 $\pm$ 0.014 & 0.260 $\pm$ 0.011 & 0.218 $\pm$ 0.020 & 0.191 $\pm$ 0.019 \\
          &   & BCC & 0.678 $\pm$ 0.008 & 0.422 $\pm$ 0.009 & 0.249 $\pm$ 0.006 & 0.212 $\pm$ 0.023 & 0.187 $\pm$ 0.021 \\
          &   & PE & 0.674 $\pm$ 0.012 & 0.422 $\pm$ 0.009 & 0.262 $\pm$ 0.012 & 0.219 $\pm$ 0.017 & 0.194 $\pm$ 0.022 \\
          &   & EA & 0.674 $\pm$ 0.011 & 0.422 $\pm$ 0.009 & 0.261 $\pm$ 0.013 & 0.218 $\pm$ 0.017 & 0.193 $\pm$ 0.021 \\
          &   & SDS & \textbf{0.689} $\pm$ 0.010 & \textbf{0.437} $\pm$ 0.010 & \textbf{0.268} $\pm$ 0.011 & \textbf{0.222} $\pm$ 0.016 & \textbf{0.195} $\pm$ 0.020 \\
        \cmidrule{2-8}
          & \multirow{5}{0.8cm}{ECE} & DS & 0.305 $\pm$ 0.012 & 0.545 $\pm$ 0.029 & 0.705 $\pm$ 0.028 & 0.739 $\pm$ 0.043 & 0.761 $\pm$ 0.046 \\
          &   & BCC & 0.302 $\pm$ 0.016 & 0.519 $\pm$ 0.072 & 0.483 $\pm$ 0.082 & \textbf{0.404} $\pm$ 0.077 & \textbf{0.369} $\pm$ 0.064 \\
          &   & PE & \textbf{0.140} $\pm$ 0.087 & 0.355 $\pm$ 0.095 & 0.493 $\pm$ 0.096 & 0.519 $\pm$ 0.097 & 0.527 $\pm$ 0.101 \\
\end{longtable}
}

\newpage
{
\scriptsize
\begin{longtable}{p{1cm}p{0.8cm}p{0.8cm}llllllll}
    \caption{Results on CIFAR10 with blur corruptions of increasing severity.}
    \label{tab:cifar10_blur}\\
        \toprule
        Corruption & Metric & Method & Severity 1 & Severity 2 & Severity 3 & Severity 4 & Severity 5 \\
        \midrule
        \endfirsthead
    \caption[]{Results on CIFAR10 with blur corruptions of increasing severity (continued).}
    \label{tab:cifar10_blur_c}\\
        \toprule
        Corruption & Metric & Method & Severity 1 & Severity 2 & Severity 3 & Severity 4 & Severity 5 \\
        \midrule
        \endhead
        \midrule
        \multicolumn{9}{r}{\textit{Continued on next page}}\\
        \endfoot
        \bottomrule
        \endlastfoot
        \multirow{20}{1cm}{Defocus Blur} & \multirow{5}{0.8cm}{Accuracy} & DS & 0.943 $\pm$ 0.006 & 0.926 $\pm$ 0.003 & 0.870 $\pm$ 0.004 & 0.755 $\pm$ 0.013 & 0.436 $\pm$ 0.016 \\
          &   & BCC & 0.944 $\pm$ 0.005 & 0.926 $\pm$ 0.003 & 0.870 $\pm$ 0.004 & 0.758 $\pm$ 0.015 & 0.398 $\pm$ 0.072 \\
          &   & PE & 0.944 $\pm$ 0.004 & 0.930 $\pm$ 0.001 & 0.878 $\pm$ 0.002 & 0.766 $\pm$ 0.012 & 0.451 $\pm$ 0.015 \\
          &   & EA & 0.945 $\pm$ 0.004 & 0.930 $\pm$ 0.001 & 0.877 $\pm$ 0.002 & 0.766 $\pm$ 0.014 & 0.450 $\pm$ 0.017 \\
          &   & SDS & \textbf{0.946} $\pm$ 0.003 & \textbf{0.932} $\pm$ \textbf{0.001} & \textbf{0.882} $\pm$ \textbf{0.002} & \textbf{0.777} $\pm$ 0.012 & \textbf{0.465} $\pm$ 0.016 \\
        \cmidrule{2-8}
          & \multirow{5}{0.8cm}{ECE} & DS & 0.054 $\pm$ 0.003 & 0.069 $\pm$ 0.002 & 0.118 $\pm$ 0.007 & 0.216 $\pm$ 0.023 & 0.458 $\pm$ 0.069 \\
          &   & BCC & \textbf{0.052} $\pm$ 0.002 & \textbf{0.067} $\pm$ 0.004 & 0.115 $\pm$ 0.011 & 0.211 $\pm$ 0.030 & 0.503 $\pm$ 0.016 \\
          &   & PE & 0.074 $\pm$ 0.097 & 0.082 $\pm$ 0.101 & 0.087 $\pm$ 0.085 & 0.100 $\pm$ 0.025 & 0.200 $\pm$ 0.093 \\
\end{longtable}
}

{
\scriptsize
\begin{longtable}{p{1cm}p{0.8cm}p{0.8cm}llllllll}
    \caption{Results on CIFAR10 with weather corruptions of increasing severity.}
    \label{tab:cifar10_weather}\\
        \toprule
        Corruption & Metric & Method & Severity 1 & Severity 2 & Severity 3 & Severity 4 & Severity 5 \\
        \midrule
        \endfirsthead
    \caption[]{Results on CIFAR10 with weather corruptions of increasing severity (continued).}
    \label{tab:cifar10_weather_c}\\
        \toprule
        Corruption & Metric & Method & Severity 1 & Severity 2 & Severity 3 & Severity 4 & Severity 5 \\
        \midrule
        \endhead
        \midrule
        \multicolumn{9}{r}{\textit{Continued on next page}}\\
        \endfoot
        \bottomrule
        \endlastfoot
        \multirow{20}{1cm}{Snow} & \multirow{5}{0.8cm}{Accuracy} & DS & 0.887 $\pm$ 0.013 & 0.767 $\pm$ 0.024 & 0.809 $\pm$ 0.019 & 0.782 $\pm$ 0.018 & 0.726 $\pm$ 0.018 \\
          &   & BCC & 0.887 $\pm$ 0.012 & 0.769 $\pm$ 0.021 & 0.811 $\pm$ 0.015 & 0.784 $\pm$ 0.015 & 0.728 $\pm$ 0.014 \\
          &   & PE & 0.889 $\pm$ 0.012 & 0.770 $\pm$ 0.023 & 0.810 $\pm$ 0.021 & 0.785 $\pm$ 0.018 & 0.728 $\pm$ 0.021 \\
          &   & EA & 0.890 $\pm$ 0.012 & 0.771 $\pm$ 0.023 & 0.810 $\pm$ 0.021 & 0.787 $\pm$ 0.016 & 0.729 $\pm$ 0.020 \\
          &   & SDS & \textbf{0.892} $\pm$ 0.010 & \textbf{0.780} $\pm$ 0.020 & \textbf{0.817} $\pm$ 0.016 & \textbf{0.794} $\pm$ 0.014 & \textbf{0.739} $\pm$ 0.018 \\
        \cmidrule{2-8}
          & \multirow{5}{0.8cm}{ECE} & DS & 0.106 $\pm$ 0.008 & 0.213 $\pm$ 0.017 & 0.176 $\pm$ 0.011 & 0.200 $\pm$ 0.009 & 0.249 $\pm$ 0.009 \\
          &   & BCC & 0.103 $\pm$ 0.005 & 0.208 $\pm$ 0.013 & 0.173 $\pm$ 0.009 & 0.197 $\pm$ 0.008 & 0.247 $\pm$ 0.010 \\
          &   & PE & \textbf{0.074} $\pm$ 0.062 & \textbf{0.096} $\pm$ 0.015 & \textbf{0.087} $\pm$ 0.021 & \textbf{0.094} $\pm$ 0.009 & \textbf{0.110} $\pm$ 0.027 \\
\end{longtable}
}

{
\scriptsize
\begin{longtable}{p{1cm}p{0.8cm}p{0.8cm}llllllll}
    \caption{Results on CIFAR10 with digital corruptions of increasing severity.}
    \label{tab:cifar10_digital}\\
        \toprule
        Corruption & Metric & Method & Severity 1 & Severity 2 & Severity 3 & Severity 4 & Severity 5 \\
        \midrule
        \endfirsthead
    \caption[]{Results on CIFAR10 with digital corruptions of increasing severity (continued).}
    \label{tab:cifar10_digital_c}\\
        \toprule
        Corruption & Metric & Method & Severity 1 & Severity 2 & Severity 3 & Severity 4 & Severity 5 \\
        \midrule
        \endhead
        \midrule
        \multicolumn{9}{r}{\textit{Continued on next page}}\\
        \endfoot
        \bottomrule
        \endlastfoot
        \multirow{20}{1cm}{Contrast} & \multirow{5}{0.8cm}{Accuracy} & DS & 0.920 $\pm$ 0.005 & 0.839 $\pm$ 0.008 & 0.749 $\pm$ 0.012 & 0.588 $\pm$ 0.033 & 0.200 $\pm$ 0.021 \\
          &   & BCC & 0.921 $\pm$ 0.004 & 0.842 $\pm$ 0.011 & 0.754 $\pm$ 0.018 & 0.548 $\pm$ 0.046 & 0.181 $\pm$ 0.041 \\
          &   & PE & 0.923 $\pm$ 0.004 & 0.840 $\pm$ 0.008 & 0.752 $\pm$ 0.012 & 0.583 $\pm$ 0.024 & 0.207 $\pm$ 0.029 \\
          &   & EA & 0.924 $\pm$ 0.004 & 0.840 $\pm$ 0.008 & 0.753 $\pm$ 0.014 & 0.585 $\pm$ 0.024 & 0.206 $\pm$ 0.029 \\
          &   & SDS & \textbf{0.926} $\pm$ 0.003 & \textbf{0.847} $\pm$ 0.008 & \textbf{0.764} $\pm$ 0.015 & \textbf{0.603} $\pm$ 0.027 & \textbf{0.208} $\pm$ 0.030 \\
        \cmidrule{2-8}
          & \multirow{5}{0.8cm}{ECE} & DS & 0.075 $\pm$ 0.002 & 0.147 $\pm$ 0.015 & 0.226 $\pm$ 0.029 & 0.369 $\pm$ 0.064 & 0.636 $\pm$ 0.205 \\
          &   & BCC & \textbf{0.072} $\pm$ 0.003 & 0.141 $\pm$ 0.023 & 0.220 $\pm$ 0.037 & 0.408 $\pm$ 0.023 & 0.463 $\pm$ 0.037 \\
          &   & PE & 0.083 $\pm$ 0.095 & \textbf{0.099} $\pm$ 0.061 & \textbf{0.117} $\pm$ 0.017 & 0.169 $\pm$ 0.092 & 0.442 $\pm$ 0.209 \\
\end{longtable}
}

\subsection{CIFAR100}
\label{sec:all_results_cifar100}
Tables~\ref{tab:cifar100_noise}--\ref{tab:cifar100_digital} present results for all corruptions on CIFAR100. %Accuracy and NLL results are similar for both methods, whereas SDS outperforms EA in Brier score and ECE. On only JPEG and Snow corruptions ECE for SDS is higher for some severity level than for EA. However, absolute values for these two corruptions are very small for both methods.

{
\scriptsize
\begin{longtable}{p{1cm}p{0.8cm}p{0.8cm}llllllll}
    \caption{Results on CIFAR100 with noise corruptions of increasing severity.}
    \label{tab:cifar100_noise}\\
        \toprule
        Corruption & Metric & Method & Severity 1 & Severity 2 & Severity 3 & Severity 4 & Severity 5 \\
        \midrule
        \endfirsthead
    \caption[]{Results on CIFAR100 with noise corruptions of increasing severity (continued).}\\
    \label{tab:cifar100_noise_c}\\
        \toprule
        Corruption & Metric & Method & Severity 1 & Severity 2 & Severity 3 & Severity 4 & Severity 5 \\
        \midrule
        \endhead
        \midrule
        \multicolumn{9}{r}{\textit{Continued on next page}}\\
        \endfoot
        \bottomrule
        \endlastfoot
        \multirow{20}{1cm}{Gaussian Noise} & \multirow{5}{0.8cm}{Accuracy} & DS & 0.418 $\pm$ 0.010 & 0.217 $\pm$ 0.018 & 0.118 $\pm$ 0.015 & 0.088 $\pm$ 0.013 & 0.072 $\pm$ 0.011 \\
          &   & BCC & 0.131 $\pm$ 0.019 & 0.035 $\pm$ 0.009 & 0.019 $\pm$ 0.007 & 0.012 $\pm$ 0.001 & 0.012 $\pm$ 0.003 \\
          &   & PE & \textbf{0.427} $\pm$ 0.012 & \textbf{0.225} $\pm$ 0.023 & \textbf{0.124} $\pm$ 0.020 & \textbf{0.092} $\pm$ 0.018 & \textbf{0.073} $\pm$ 0.017 \\
          &   & EA & 0.418 $\pm$ 0.012 & 0.215 $\pm$ 0.018 & 0.116 $\pm$ 0.015 & 0.086 $\pm$ 0.015 & 0.069 $\pm$ 0.014 \\
          &   & SDS & 0.419 $\pm$ 0.013 & 0.213 $\pm$ 0.017 & 0.116 $\pm$ 0.015 & 0.086 $\pm$ 0.015 & 0.069 $\pm$ 0.014 \\
        \cmidrule{2-8}
          & \multirow{5}{0.8cm}{ECE} & DS & 0.533 $\pm$ 0.008 & 0.708 $\pm$ 0.007 & 0.789 $\pm$ 0.010 & 0.822 $\pm$ 0.010 & 0.829 $\pm$ 0.013 \\
          &   & BCC & 0.811 $\pm$ 0.021 & 0.904 $\pm$ 0.027 & 0.929 $\pm$ 0.033 & 0.943 $\pm$ 0.028 & 0.944 $\pm$ 0.035 \\
          &   & PE & 0.103 $\pm$ 0.018 & \textbf{0.110} $\pm$ \textbf{0.044} & \textbf{0.218} $\pm$ \textbf{0.058} & \textbf{0.251} $\pm$ \textbf{0.063} & \textbf{0.265} $\pm$ \textbf{0.066} \\
\end{longtable}
}

\newpage
{
\scriptsize
\begin{longtable}{p{1cm}p{0.8cm}p{0.8cm}llllllll}
    \caption{Results on CIFAR100 with blur corruptions of increasing severity.}
    \label{tab:cifar100_blur}\\
        \toprule
        Corruption & Metric & Method & Severity 1 & Severity 2 & Severity 3 & Severity 4 & Severity 5 \\
        \midrule
        \endfirsthead
    \caption[]{Results on CIFAR100 with blur corruptions of increasing severity (continued).}\\
    \label{tab:cifar100_blur_c}\\
        \toprule
        Corruption & Metric & Method & Severity 1 & Severity 2 & Severity 3 & Severity 4 & Severity 5 \\
        \midrule
        \endhead
        \midrule
        \multicolumn{9}{r}{\textit{Continued on next page}}\\
        \endfoot
        \bottomrule
        \endlastfoot
        \multirow{20}{1cm}{Defocus Blur} & \multirow{5}{0.8cm}{Accuracy} & DS & 0.805 $\pm$ 0.002 & 0.766 $\pm$ 0.003 & 0.668 $\pm$ 0.006 & 0.524 $\pm$ 0.004 & 0.288 $\pm$ 0.004 \\
          &   & BCC & 0.684 $\pm$ 0.004 & 0.573 $\pm$ 0.020 & 0.397 $\pm$ 0.002 & 0.219 $\pm$ 0.015 & 0.092 $\pm$ 0.006 \\
          &   & PE & \textbf{0.813} $\pm$ 0.002 & 0.777 $\pm$ 0.004 & \textbf{0.682} $\pm$ \textbf{0.003} & \textbf{0.539} $\pm$ \textbf{0.005} & \textbf{0.300} $\pm$ \textbf{0.007} \\
          &   & EA & 0.810 $\pm$ 0.002 & 0.771 $\pm$ 0.003 & 0.671 $\pm$ 0.003 & 0.524 $\pm$ 0.003 & 0.285 $\pm$ 0.002 \\
          &   & SDS & \textbf{0.813} $\pm$ 0.001 & \textbf{0.778} $\pm$ 0.002 & 0.677 $\pm$ 0.000 & 0.523 $\pm$ 0.005 & 0.283 $\pm$ 0.001 \\
        \cmidrule{2-8}
          & \multirow{5}{0.8cm}{ECE} & DS & 0.192 $\pm$ 0.002 & 0.228 $\pm$ 0.003 & 0.319 $\pm$ 0.007 & 0.454 $\pm$ 0.006 & 0.655 $\pm$ 0.022 \\
          &   & BCC & 0.296 $\pm$ 0.004 & 0.405 $\pm$ 0.021 & 0.571 $\pm$ 0.004 & 0.735 $\pm$ 0.014 & 0.847 $\pm$ 0.002 \\
          &   & PE & 0.243 $\pm$ 0.005 & 0.245 $\pm$ 0.005 & 0.232 $\pm$ 0.004 & 0.168 $\pm$ 0.007 & \textbf{0.076} $\pm$ \textbf{0.001} \\
\end{longtable}
}

{
\scriptsize
\begin{longtable}{p{1cm}p{0.8cm}p{0.8cm}llllllll}
    \caption{Results on CIFAR100 with weather corruptions of increasing severity.}
    \label{tab:cifar100_weather}\\
        \toprule
        Corruption & Metric & Method & Severity 1 & Severity 2 & Severity 3 & Severity 4 & Severity 5 \\
        \midrule
        \endfirsthead
    \caption[]{Results on CIFAR100 with weather corruptions of increasing severity (continued).}\\
    \label{tab:cifar100_weather}\\
        \toprule
        Corruption & Metric & Method & Severity 1 & Severity 2 & Severity 3 & Severity 4 & Severity 5 \\
        \midrule
        \endhead
        \midrule
        \multicolumn{9}{r}{\textit{Continued on next page}}\\
        \endfoot
        \bottomrule
        \endlastfoot
        \multirow{20}{1cm}{Snow} & \multirow{5}{0.8cm}{Accuracy} & DS & 0.699 $\pm$ 0.003 & 0.538 $\pm$ 0.005 & 0.606 $\pm$ 0.007 & 0.590 $\pm$ 0.003 & 0.508 $\pm$ 0.004 \\
          &   & BCC & 0.462 $\pm$ 0.024 & 0.237 $\pm$ 0.013 & 0.302 $\pm$ 0.011 & 0.308 $\pm$ 0.012 & 0.206 $\pm$ 0.012 \\
          &   & PE & 0.711 $\pm$ 0.001 & \textbf{0.552} $\pm$ 0.007 & \textbf{0.618} $\pm$ 0.004 & \textbf{0.602} $\pm$ 0.006 & \textbf{0.516} $\pm$ 0.005 \\
          &   & EA & 0.706 $\pm$ 0.006 & 0.546 $\pm$ 0.004 & 0.615 $\pm$ 0.005 & 0.595 $\pm$ 0.004 & 0.511 $\pm$ 0.006 \\
          &   & SDS & \textbf{0.713} $\pm$ 0.007 & 0.548 $\pm$ 0.009 & \textbf{0.618} $\pm$ 0.007 & 0.599 $\pm$ 0.004 & 0.513 $\pm$ 0.004 \\
        \cmidrule{2-8}
          & \multirow{5}{0.8cm}{ECE} & DS & 0.292 $\pm$ 0.003 & 0.438 $\pm$ 0.004 & 0.379 $\pm$ 0.007 & 0.395 $\pm$ 0.004 & 0.468 $\pm$ 0.003 \\
          &   & BCC & 0.506 $\pm$ 0.026 & 0.718 $\pm$ 0.013 & 0.655 $\pm$ 0.011 & 0.643 $\pm$ 0.012 & 0.742 $\pm$ 0.013 \\
          &   & PE & 0.261 $\pm$ 0.003 & 0.202 $\pm$ 0.004 & 0.242 $\pm$ 0.003 & 0.238 $\pm$ 0.007 & 0.194 $\pm$ 0.007 \\
\end{longtable}
}

{
\scriptsize
\begin{longtable}{p{1cm}p{0.8cm}p{0.8cm}llllllll}
    \caption{Results on CIFAR100 with digital corruptions of increasing severity.}
    \label{tab:cifar100_digital}\\
        \toprule
        Corruption & Metric & Method & Severity 1 & Severity 2 & Severity 3 & Severity 4 & Severity 5 \\
        \midrule
        \endfirsthead
    \caption[]{Results on CIFAR100 with digital corruptions of increasing severity (continued).}\\
        \toprule
        Corruption & Metric & Method & Severity 1 & Severity 2 & Severity 3 & Severity 4 & Severity 5 \\
        \midrule
        \endhead
        \midrule
        \multicolumn{9}{r}{\textit{Continued on next page}}\\
        \endfoot
        \bottomrule
        \endlastfoot
        \multirow{20}{1cm}{Contrast} & \multirow{5}{0.8cm}{Accuracy} & DS & 0.779 $\pm$ 0.003 & 0.684 $\pm$ 0.000 & 0.596 $\pm$ 0.001 & 0.454 $\pm$ 0.008 & 0.150 $\pm$ 0.007 \\
          &   & BCC & 0.613 $\pm$ 0.005 & 0.419 $\pm$ 0.013 & 0.276 $\pm$ 0.014 & 0.147 $\pm$ 0.028 & 0.039 $\pm$ 0.005 \\
          &   & PE & \textbf{0.791} $\pm$ 0.002 & \textbf{0.707} $\pm$ \textbf{0.001} & \textbf{0.621} $\pm$ \textbf{0.001} & \textbf{0.481} $\pm$ \textbf{0.003} & \textbf{0.162} $\pm$ \textbf{0.003} \\
          &   & EA & 0.783 $\pm$ 0.002 & 0.689 $\pm$ 0.002 & 0.597 $\pm$ 0.002 & 0.453 $\pm$ 0.006 & 0.145 $\pm$ 0.006 \\
          &   & SDS & 0.790 $\pm$ 0.003 & 0.699 $\pm$ 0.003 & 0.608 $\pm$ 0.002 & 0.451 $\pm$ 0.005 & 0.144 $\pm$ 0.006 \\
        \cmidrule{2-8}
          & \multirow{5}{0.8cm}{ECE} & DS & 0.216 $\pm$ 0.002 & 0.307 $\pm$ 0.001 & 0.386 $\pm$ 0.003 & 0.511 $\pm$ 0.010 & 0.784 $\pm$ 0.012 \\
          &   & BCC & 0.366 $\pm$ 0.005 & 0.550 $\pm$ 0.014 & 0.688 $\pm$ 0.014 & 0.804 $\pm$ 0.029 & 0.898 $\pm$ 0.007 \\
          &   & PE & 0.246 $\pm$ 0.006 & 0.244 $\pm$ 0.004 & 0.212 $\pm$ 0.002 & 0.133 $\pm$ 0.006 & \textbf{0.171} $\pm$ \textbf{0.017} \\
\end{longtable}
}

\subsection{ImageNet}
\label{sec:full_results_imagenet}
Results for all corruptions on ImageNet are provided in~Tables~\ref{tab:imagenet_noise}--\ref{tab:imagenet_digital}. On some corruptions BCC fails to provide reasonable predictions and outputs uniform probabilities for all classes. In these cases ECE is not adequately computed as being very close to zero. Please note that all these cases paired with very low accuracy. See, for example, ``Shot Noise'' corruption of severity 2. ECE values for such cases are non-informative.  %For all corruptions SDS and EA have similar accuracy, Brier score and NLL. In terms of ECE, for most of corruption types SDS demonstrates better results than EA.

{
\scriptsize
\begin{longtable}{p{1cm}p{0.8cm}p{0.8cm}llllllll}
    \caption{Results on ImageNet with noise corruptions of increasing severity.}
    \label{tab:imagenet_noise}\\
        \toprule
        Corruption & Metric & Method & Severity 1 & Severity 2 & Severity 3 & Severity 4 & Severity 5 \\
        \midrule
        \endfirsthead
    \caption[]{Results on ImageNet with noise corruptions of increasing severity (continued).}\\
        \toprule
        Corruption & Metric & Method & Severity 1 & Severity 2 & Severity 3 & Severity 4 & Severity 5 \\
        \midrule
        \endhead
        \midrule
        \multicolumn{9}{r}{\textit{Continued on next page}}\\
        \endfoot
        \bottomrule
        \endlastfoot
        \multirow{16}{1cm}{Gaussian Noise} & \multirow{4}{0.8cm}{Accuracy} & DS & 0.630 $\pm$ 0.003 & 0.535 $\pm$ 0.002 & 0.374 $\pm$ 0.004 & 0.198 $\pm$ 0.005 & \textbf{0.064} $\pm$ 0.004 \\
          &   & BCC & 0.613 $\pm$ 0.002 & 0.153 $\pm$ 0.261 & 0.002 $\pm$ 0.001 & 0.001 $\pm$ 0.000 & 0.001 $\pm$ 0.000 \\
          &   & EA & \textbf{0.638} $\pm$ \textbf{0.001} & \textbf{0.541} $\pm$ \textbf{0.001} & \textbf{0.379} $\pm$ 0.004 & \textbf{0.199} $\pm$ 0.004 & 0.063 $\pm$ 0.005 \\
          &   & SDS & \textbf{0.638} $\pm$ \textbf{0.001} & \textbf{0.541} $\pm$ \textbf{0.001} & \textbf{0.379} $\pm$ 0.004 & \textbf{0.199} $\pm$ 0.004 & 0.063 $\pm$ 0.005 \\
        \cmidrule{2-8}
          & \multirow{4}{0.8cm}{ECE} & DS & 0.365 $\pm$ 0.003 & 0.457 $\pm$ 0.002 & 0.606 $\pm$ 0.002 & 0.733 $\pm$ 0.003 & 0.637 $\pm$ 0.009 \\
          &   & BCC & 0.612 $\pm$ 0.002 & 0.152 $\pm$ 0.261 & 0.924 $\pm$ 0.019 & 0.925 $\pm$ 0.015 & 0.938 $\pm$ 0.005 \\
          &   & EA & \textbf{0.022} $\pm$ \textbf{0.001} & \textbf{0.022} $\pm$ 0.001 & 0.029 $\pm$ 0.004 & 0.071 $\pm$ 0.003 & 0.116 $\pm$ 0.009 \\
          &   & SDS & 0.025 $\pm$ 0.001 & 0.023 $\pm$ 0.001 & \textbf{0.027} $\pm$ 0.003 & \textbf{0.066} $\pm$ 0.003 & \textbf{0.113} $\pm$ 0.009 \\
        \cmidrule{2-8}
          & \multirow{4}{0.8cm}{Brier score} & DS & 0.733 $\pm$ 0.005 & 0.919 $\pm$ 0.004 & 1.223 $\pm$ 0.004 & 1.505 $\pm$ 0.006 & 1.475 $\pm$ 0.008 \\
          &   & BCC & 0.999 $\pm$ 0.000 & 0.999 $\pm$ 0.000 & 1.885 $\pm$ 0.024 & 1.890 $\pm$ 0.023 & 1.910 $\pm$ 0.007 \\
          &   & EA & \textbf{0.481} $\pm$ \textbf{0.002} & \textbf{0.588} $\pm$ \textbf{0.002} & \textbf{0.751} $\pm$ \textbf{0.002} & 0.910 $\pm$ 0.003 & 1.006 $\pm$ 0.005 \\
          &   & SDS & \textbf{0.481} $\pm$ \textbf{0.002} & \textbf{0.588} $\pm$ \textbf{0.002} & \textbf{0.751} $\pm$ \textbf{0.002} & \textbf{0.909} $\pm$ 0.003 & \textbf{1.004} $\pm$ 0.005 \\
        \cmidrule{2-8}
          & \multirow{4}{0.8cm}{NLL} & DS & 9.433 $\pm$ 0.039 & 12.010 $\pm$ 0.024 & 16.295 $\pm$ 0.106 & 20.573 $\pm$ 0.135 & 23.449 $\pm$ 0.177 \\
          &   & BCC & 6.907 $\pm$ 0.000 & 6.907 $\pm$ 0.000 & 11.917 $\pm$ 0.678 & 13.904 $\pm$ 0.474 & 17.160 $\pm$ 0.489 \\
          &   & EA & \textbf{1.542} $\pm$ 0.010 & \textbf{2.062} $\pm$ 0.010 & \textbf{3.065} $\pm$ 0.050 & \textbf{4.510} $\pm$ \textbf{0.102} & 6.197 $\pm$ 0.093 \\
          &   & SDS & 1.554 $\pm$ 0.011 & 2.079 $\pm$ 0.009 & 3.081 $\pm$ 0.050 & \textbf{4.510} $\pm$ \textbf{0.097} & \textbf{6.147} $\pm$ 0.080 \\
        \midrule
        \multirow{16}{1cm}{Shot Noise} & \multirow{4}{0.8cm}{Accuracy} & DS & 0.618 $\pm$ 0.003 & 0.506 $\pm$ 0.005 & 0.358 $\pm$ 0.001 & \textbf{0.159} $\pm$ 0.002 & \textbf{0.074} $\pm$ 0.004 \\
          &   & BCC & 0.598 $\pm$ 0.003 & 0.002 $\pm$ 0.001 & 0.002 $\pm$ 0.001 & 0.001 $\pm$ 0.001 & 0.001 $\pm$ 0.000 \\
          &   & EA & \textbf{0.625} $\pm$ \textbf{0.003} & \textbf{0.512} $\pm$ 0.004 & \textbf{0.362} $\pm$ \textbf{0.003} & 0.158 $\pm$ 0.002 & 0.072 $\pm$ 0.004 \\
          &   & SDS & \textbf{0.625} $\pm$ \textbf{0.003} & \textbf{0.512} $\pm$ 0.004 & \textbf{0.362} $\pm$ \textbf{0.003} & 0.158 $\pm$ 0.002 & 0.072 $\pm$ 0.004 \\
        \cmidrule{2-8}
          & \multirow{4}{0.8cm}{ECE} & DS & 0.377 $\pm$ 0.003 & 0.485 $\pm$ 0.004 & 0.619 $\pm$ 0.002 & 0.737 $\pm$ 0.004 & 0.638 $\pm$ 0.008 \\
          &   & BCC & 0.597 $\pm$ 0.003 & 0.001 $\pm$ 0.001 & 0.931 $\pm$ 0.022 & 0.925 $\pm$ 0.018 & 0.940 $\pm$ 0.006 \\
          &   & EA & \textbf{0.021} $\pm$ \textbf{0.001} & \textbf{0.021} $\pm$ 0.002 & 0.034 $\pm$ 0.004 & 0.091 $\pm$ 0.004 & 0.123 $\pm$ 0.004 \\
          &   & SDS & 0.024 $\pm$ 0.001 & 0.022 $\pm$ 0.003 & \textbf{0.031} $\pm$ 0.004 & \textbf{0.086} $\pm$ 0.004 & \textbf{0.119} $\pm$ 0.004 \\
        \cmidrule{2-8}
          & \multirow{4}{0.8cm}{Brier score} & DS & 0.756 $\pm$ 0.007 & 0.974 $\pm$ 0.009 & 1.250 $\pm$ 0.003 & 1.534 $\pm$ 0.004 & 1.469 $\pm$ 0.005 \\
          &   & BCC & 0.999 $\pm$ 0.000 & 0.999 $\pm$ 0.000 & 1.896 $\pm$ 0.029 & 1.890 $\pm$ 0.027 & 1.912 $\pm$ 0.009 \\
          &   & EA & \textbf{0.495} $\pm$ \textbf{0.004} & \textbf{0.619} $\pm$ \textbf{0.005} & 0.768 $\pm$ 0.004 & 0.943 $\pm$ 0.001 & 1.003 $\pm$ 0.002 \\
          &   & SDS & \textbf{0.495} $\pm$ \textbf{0.004} & \textbf{0.619} $\pm$ \textbf{0.005} & \textbf{0.767} $\pm$ 0.004 & \textbf{0.942} $\pm$ 0.001 & \textbf{1.002} $\pm$ 0.002 \\
        \cmidrule{2-8}
          & \multirow{4}{0.8cm}{NLL} & DS & 9.753 $\pm$ 0.098 & 12.759 $\pm$ 0.099 & 16.660 $\pm$ 0.062 & 21.280 $\pm$ 0.075 & 23.062 $\pm$ 0.148 \\
          &   & BCC & 6.907 $\pm$ 0.000 & 6.907 $\pm$ 0.000 & 12.338 $\pm$ 0.892 & 14.802 $\pm$ 0.087 & 16.841 $\pm$ 0.389 \\
          &   & EA & \textbf{1.603} $\pm$ 0.016 & \textbf{2.225} $\pm$ 0.026 & \textbf{3.190} $\pm$ 0.024 & 4.966 $\pm$ 0.066 & 6.113 $\pm$ 0.084 \\
          &   & SDS & 1.618 $\pm$ 0.019 & 2.242 $\pm$ 0.026 & 3.203 $\pm$ 0.022 & \textbf{4.950} $\pm$ 0.059 & \textbf{6.061} $\pm$ 0.066 \\
        \midrule
        \multirow{16}{1cm}{Impulse Noise} & \multirow{4}{0.8cm}{Accuracy} & DS & 0.538 $\pm$ 0.009 & 0.445 $\pm$ 0.006 & 0.356 $\pm$ 0.004 & \textbf{0.178} $\pm$ 0.008 & \textbf{0.065} $\pm$ 0.008 \\
          &   & BCC & 0.002 $\pm$ 0.000 & 0.003 $\pm$ 0.004 & 0.002 $\pm$ 0.002 & 0.001 $\pm$ 0.000 & 0.001 $\pm$ 0.000 \\
          &   & EA & 0.546 $\pm$ 0.009 & \textbf{0.452} $\pm$ 0.006 & \textbf{0.360} $\pm$ 0.004 & 0.177 $\pm$ 0.008 & 0.064 $\pm$ 0.008 \\
          &   & SDS & \textbf{0.547} $\pm$ 0.009 & \textbf{0.452} $\pm$ 0.006 & \textbf{0.360} $\pm$ 0.004 & 0.177 $\pm$ 0.008 & 0.064 $\pm$ 0.008 \\
        \cmidrule{2-8}
          & \multirow{4}{0.8cm}{ECE} & DS & 0.453 $\pm$ 0.008 & 0.541 $\pm$ 0.005 & 0.621 $\pm$ 0.005 & 0.741 $\pm$ 0.005 & 0.664 $\pm$ 0.052 \\
          &   & BCC & 0.001 $\pm$ 0.000 & 0.833 $\pm$ 0.126 & 0.929 $\pm$ 0.022 & 0.940 $\pm$ 0.006 & 0.934 $\pm$ 0.054 \\
          &   & EA & \textbf{0.023} $\pm$ 0.002 & \textbf{0.024} $\pm$ \textbf{0.002} & 0.033 $\pm$ 0.003 & 0.079 $\pm$ 0.003 & 0.119 $\pm$ 0.008 \\
          &   & SDS & 0.027 $\pm$ 0.004 & \textbf{0.024} $\pm$ \textbf{0.003} & \textbf{0.031} $\pm$ 0.003 & \textbf{0.075} $\pm$ 0.003 & \textbf{0.115} $\pm$ 0.008 \\
        \cmidrule{2-8}
          & \multirow{4}{0.8cm}{Brier score} & DS & 0.911 $\pm$ 0.017 & 1.090 $\pm$ 0.010 & 1.255 $\pm$ 0.010 & 1.528 $\pm$ 0.003 & 1.506 $\pm$ 0.057 \\
          &   & BCC & 0.999 $\pm$ 0.000 & 1.779 $\pm$ 0.138 & 1.892 $\pm$ 0.029 & 1.913 $\pm$ 0.010 & 1.904 $\pm$ 0.077 \\
          &   & EA & \textbf{0.581} $\pm$ \textbf{0.009} & \textbf{0.680} $\pm$ \textbf{0.006} & \textbf{0.769} $\pm$ \textbf{0.004} & 0.927 $\pm$ 0.006 & 1.007 $\pm$ 0.007 \\
          &   & SDS & \textbf{0.581} $\pm$ \textbf{0.009} & \textbf{0.680} $\pm$ \textbf{0.006} & \textbf{0.769} $\pm$ \textbf{0.004} & \textbf{0.926} $\pm$ 0.006 & \textbf{1.005} $\pm$ 0.006 \\
        \cmidrule{2-8}
          & \multirow{4}{0.8cm}{NLL} & DS & 11.892 $\pm$ 0.214 & 14.387 $\pm$ 0.200 & 16.753 $\pm$ 0.225 & 20.996 $\pm$ 0.072 & 23.438 $\pm$ 0.165 \\
          &   & BCC & 6.907 $\pm$ 0.000 & 10.731 $\pm$ 1.120 & 11.892 $\pm$ 0.506 & 13.938 $\pm$ 0.680 & 16.901 $\pm$ 0.699 \\
          &   & EA & \textbf{2.019} $\pm$ 0.044 & \textbf{2.575} $\pm$ 0.034 & \textbf{3.180} $\pm$ 0.036 & 4.704 $\pm$ 0.114 & 6.159 $\pm$ 0.152 \\
          &   & SDS & 2.031 $\pm$ 0.042 & 2.588 $\pm$ 0.035 & 3.195 $\pm$ 0.036 & \textbf{4.700} $\pm$ 0.109 & \textbf{6.104} $\pm$ 0.133 \\
\end{longtable}
}

\newpage
{
\scriptsize
\begin{longtable}{p{1cm}p{0.8cm}p{0.8cm}llllllll}
    \caption{Results on ImageNet with blur corruptions of increasing severity.}
    \label{tab:imagenet_blur}\\
        \toprule
        Corruption & Metric & Method & Severity 1 & Severity 2 & Severity 3 & Severity 4 & Severity 5 \\
        \midrule
        \endfirsthead
    \caption[]{Results on ImageNet with blur corruptions of increasing severity (continued).}\\
        \toprule
        Corruption & Metric & Method & Severity 1 & Severity 2 & Severity 3 & Severity 4 & Severity 5 \\
        \midrule
        \endhead
        \midrule
        \multicolumn{9}{r}{\textit{Continued on next page}}\\
        \endfoot
        \bottomrule
        \endlastfoot
        \multirow{16}{1cm}{Defocus Blur} & \multirow{4}{0.8cm}{Accuracy} & DS & 0.632 $\pm$ 0.002 & 0.565 $\pm$ 0.003 & 0.430 $\pm$ 0.007 & 0.309 $\pm$ 0.007 & 0.215 $\pm$ 0.009 \\
          &   & BCC & 0.618 $\pm$ 0.002 & 0.548 $\pm$ 0.003 & 0.001 $\pm$ 0.000 & 0.002 $\pm$ 0.000 & 0.002 $\pm$ 0.000 \\
          &   & EA & \textbf{0.638} $\pm$ \textbf{0.003} & \textbf{0.573} $\pm$ \textbf{0.003} & \textbf{0.436} $\pm$ 0.008 & \textbf{0.312} $\pm$ 0.009 & 0.215 $\pm$ 0.009 \\
          &   & SDS & \textbf{0.638} $\pm$ \textbf{0.002} & \textbf{0.573} $\pm$ \textbf{0.003} & \textbf{0.436} $\pm$ 0.008 & \textbf{0.312} $\pm$ 0.009 & 0.215 $\pm$ 0.009 \\
        \cmidrule{2-8}
          & \multirow{4}{0.8cm}{ECE} & DS & 0.364 $\pm$ 0.002 & 0.430 $\pm$ 0.002 & 0.562 $\pm$ 0.007 & 0.677 $\pm$ 0.007 & 0.756 $\pm$ 0.007 \\
          &   & BCC & 0.617 $\pm$ 0.002 & 0.547 $\pm$ 0.003 & 0.000 $\pm$ 0.000 & 0.908 $\pm$ 0.003 & 0.927 $\pm$ 0.002 \\
          &   & EA & \textbf{0.046} $\pm$ \textbf{0.002} & \textbf{0.058} $\pm$ \textbf{0.002} & \textbf{0.059} $\pm$ 0.004 & \textbf{0.036} $\pm$ 0.005 & \textbf{0.012} $\pm$ 0.004 \\
          &   & SDS & 0.053 $\pm$ 0.002 & 0.065 $\pm$ 0.002 & 0.064 $\pm$ 0.005 & 0.040 $\pm$ 0.005 & 0.013 $\pm$ 0.005 \\
        \cmidrule{2-8}
          & \multirow{4}{0.8cm}{Brier score} & DS & 0.730 $\pm$ 0.004 & 0.862 $\pm$ 0.005 & 1.128 $\pm$ 0.014 & 1.361 $\pm$ 0.014 & 1.527 $\pm$ 0.015 \\
          &   & BCC & 0.999 $\pm$ 0.000 & 0.999 $\pm$ 0.000 & 0.999 $\pm$ 0.000 & 1.863 $\pm$ 0.004 & 1.892 $\pm$ 0.003 \\
          &   & EA & \textbf{0.485} $\pm$ 0.002 & \textbf{0.560} $\pm$ 0.003 & \textbf{0.704} $\pm$ 0.007 & \textbf{0.814} $\pm$ 0.007 & \textbf{0.891} $\pm$ \textbf{0.006} \\
          &   & SDS & 0.486 $\pm$ 0.002 & 0.561 $\pm$ 0.003 & 0.705 $\pm$ 0.007 & 0.815 $\pm$ 0.007 & \textbf{0.891} $\pm$ \textbf{0.006} \\
        \cmidrule{2-8}
          & \multirow{4}{0.8cm}{NLL} & DS & 9.356 $\pm$ 0.071 & 11.191 $\pm$ 0.050 & 14.944 $\pm$ 0.234 & 18.308 $\pm$ 0.189 & 20.694 $\pm$ 0.215 \\
          &   & BCC & 6.907 $\pm$ 0.000 & 6.907 $\pm$ 0.000 & 6.908 $\pm$ 0.000 & 11.056 $\pm$ 0.086 & 11.953 $\pm$ 0.107 \\
          &   & EA & \textbf{1.536} $\pm$ 0.010 & \textbf{1.885} $\pm$ 0.016 & \textbf{2.671} $\pm$ 0.046 & \textbf{3.472} $\pm$ 0.071 & \textbf{4.206} $\pm$ 0.078 \\
          &   & SDS & 1.552 $\pm$ 0.009 & 1.901 $\pm$ 0.015 & 2.687 $\pm$ 0.046 & 3.489 $\pm$ 0.071 & 4.216 $\pm$ 0.074 \\
        \midrule
        \multirow{16}{1cm}{Frosted Glass Blur} & \multirow{4}{0.8cm}{Accuracy} & DS & 0.570 $\pm$ 0.003 & 0.436 $\pm$ 0.005 & \textbf{0.192} $\pm$ 0.006 & \textbf{0.147} $\pm$ 0.004 & \textbf{0.109} $\pm$ 0.003 \\
          &   & BCC & 0.001 $\pm$ 0.000 & 0.010 $\pm$ 0.001 & 0.002 $\pm$ 0.000 & 0.002 $\pm$ 0.001 & 0.002 $\pm$ 0.000 \\
          &   & EA & \textbf{0.573} $\pm$ 0.002 & 0.436 $\pm$ 0.004 & 0.187 $\pm$ 0.005 & 0.144 $\pm$ 0.004 & 0.106 $\pm$ 0.003 \\
          &   & SDS & \textbf{0.573} $\pm$ 0.002 & 0.436 $\pm$ 0.004 & 0.187 $\pm$ 0.005 & 0.144 $\pm$ 0.004 & 0.106 $\pm$ 0.003 \\
        \cmidrule{2-8}
          & \multirow{4}{0.8cm}{ECE} & DS & 0.424 $\pm$ 0.003 & 0.551 $\pm$ 0.005 & 0.738 $\pm$ 0.002 & 0.764 $\pm$ 0.003 & 0.801 $\pm$ 0.007 \\
          &   & BCC & 0.332 $\pm$ 0.187 & 0.863 $\pm$ 0.003 & 0.931 $\pm$ 0.003 & 0.937 $\pm$ 0.001 & 0.927 $\pm$ 0.006 \\
          &   & EA & \textbf{0.016} $\pm$ \textbf{0.001} & 0.014 $\pm$ 0.001 & 0.110 $\pm$ 0.006 & 0.115 $\pm$ 0.005 & 0.096 $\pm$ 0.008 \\
          &   & SDS & 0.019 $\pm$ 0.001 & \textbf{0.009} $\pm$ \textbf{0.001} & \textbf{0.104} $\pm$ 0.006 & \textbf{0.110} $\pm$ 0.005 & \textbf{0.092} $\pm$ 0.008 \\
        \cmidrule{2-8}
          & \multirow{4}{0.8cm}{Brier score} & DS & 0.851 $\pm$ 0.005 & 1.109 $\pm$ 0.009 & 1.513 $\pm$ 0.004 & 1.576 $\pm$ 0.007 & 1.650 $\pm$ 0.010 \\
          &   & BCC & 1.286 $\pm$ 0.186 & 1.808 $\pm$ 0.005 & 1.904 $\pm$ 0.004 & 1.909 $\pm$ 0.002 & 1.893 $\pm$ 0.007 \\
          &   & EA & \textbf{0.555} $\pm$ \textbf{0.003} & \textbf{0.698} $\pm$ \textbf{0.004} & 0.936 $\pm$ 0.006 & 0.969 $\pm$ 0.004 & 0.983 $\pm$ 0.004 \\
          &   & SDS & \textbf{0.555} $\pm$ \textbf{0.003} & \textbf{0.698} $\pm$ \textbf{0.004} & \textbf{0.934} $\pm$ 0.006 & \textbf{0.967} $\pm$ 0.004 & \textbf{0.982} $\pm$ 0.004 \\
        \cmidrule{2-8}
          & \multirow{4}{0.8cm}{NLL} & DS & 10.984 $\pm$ 0.084 & 14.554 $\pm$ 0.121 & 20.070 $\pm$ 0.061 & 21.252 $\pm$ 0.155 & 22.547 $\pm$ 0.182 \\
          &   & BCC & 8.249 $\pm$ 0.806 & 10.855 $\pm$ 0.060 & 13.982 $\pm$ 0.177 & 15.305 $\pm$ 0.178 & 15.228 $\pm$ 0.145 \\
          &   & EA & \textbf{1.883} $\pm$ 0.019 & \textbf{2.698} $\pm$ 0.035 & 4.702 $\pm$ 0.076 & 5.141 $\pm$ 0.055 & 5.510 $\pm$ 0.044 \\
          &   & SDS & 1.896 $\pm$ 0.019 & 2.714 $\pm$ 0.034 & \textbf{4.675} $\pm$ 0.069 & \textbf{5.102} $\pm$ 0.050 & \textbf{5.464} $\pm$ 0.043 \\
        \midrule
        \multirow{16}{1cm}{Motion Blur} & \multirow{4}{0.8cm}{Accuracy} & DS & 0.674 $\pm$ 0.003 & 0.576 $\pm$ 0.003 & 0.416 $\pm$ 0.004 & 0.253 $\pm$ 0.002 & 0.173 $\pm$ 0.003 \\
          &   & BCC & 0.664 $\pm$ 0.004 & 0.001 $\pm$ 0.000 & 0.005 $\pm$ 0.003 & 0.002 $\pm$ 0.000 & 0.001 $\pm$ 0.000 \\
          &   & EA & \textbf{0.682} $\pm$ \textbf{0.003} & 0.584 $\pm$ 0.003 & \textbf{0.423} $\pm$ \textbf{0.002} & \textbf{0.256} $\pm$ 0.003 & \textbf{0.174} $\pm$ 0.003 \\
          &   & SDS & \textbf{0.682} $\pm$ \textbf{0.003} & \textbf{0.585} $\pm$ 0.003 & \textbf{0.423} $\pm$ \textbf{0.002} & \textbf{0.256} $\pm$ 0.003 & \textbf{0.174} $\pm$ 0.003 \\
        \cmidrule{2-8}
          & \multirow{4}{0.8cm}{ECE} & DS & 0.322 $\pm$ 0.003 & 0.418 $\pm$ 0.003 & 0.567 $\pm$ 0.005 & 0.697 $\pm$ 0.002 & 0.750 $\pm$ 0.002 \\
          &   & BCC & 0.663 $\pm$ 0.004 & 0.000 $\pm$ 0.000 & 0.891 $\pm$ 0.051 & 0.930 $\pm$ 0.003 & 0.928 $\pm$ 0.004 \\
          &   & EA & \textbf{0.034} $\pm$ \textbf{0.001} & \textbf{0.034} $\pm$ \textbf{0.001} & \textbf{0.013} $\pm$ 0.002 & 0.055 $\pm$ 0.005 & 0.091 $\pm$ 0.008 \\
          &   & SDS & 0.040 $\pm$ 0.001 & 0.041 $\pm$ 0.001 & 0.015 $\pm$ 0.002 & \textbf{0.049} $\pm$ 0.005 & \textbf{0.086} $\pm$ 0.008 \\
        \cmidrule{2-8}
          & \multirow{4}{0.8cm}{Brier score} & DS & 0.646 $\pm$ 0.006 & 0.839 $\pm$ 0.006 & 1.143 $\pm$ 0.009 & 1.421 $\pm$ 0.004 & 1.542 $\pm$ 0.005 \\
          &   & BCC & 0.999 $\pm$ 0.000 & 0.999 $\pm$ 0.000 & 1.841 $\pm$ 0.067 & 1.895 $\pm$ 0.004 & 1.894 $\pm$ 0.005 \\
          &   & EA & \textbf{0.431} $\pm$ \textbf{0.003} & \textbf{0.542} $\pm$ \textbf{0.004} & \textbf{0.713} $\pm$ \textbf{0.003} & 0.871 $\pm$ 0.004 & 0.942 $\pm$ 0.006 \\
          &   & SDS & \textbf{0.431} $\pm$ \textbf{0.003} & \textbf{0.542} $\pm$ \textbf{0.004} & \textbf{0.713} $\pm$ \textbf{0.003} & \textbf{0.870} $\pm$ 0.003 & \textbf{0.941} $\pm$ 0.006 \\
        \cmidrule{2-8}
          & \multirow{4}{0.8cm}{NLL} & DS & 8.299 $\pm$ 0.070 & 11.000 $\pm$ 0.043 & 15.241 $\pm$ 0.161 & 19.245 $\pm$ 0.058 & 21.093 $\pm$ 0.090 \\
          &   & BCC & 6.907 $\pm$ 0.000 & 6.906 $\pm$ 0.002 & 11.255 $\pm$ 0.767 & 13.660 $\pm$ 0.239 & 14.951 $\pm$ 0.166 \\
          &   & EA & \textbf{1.328} $\pm$ 0.012 & \textbf{1.843} $\pm$ 0.016 & \textbf{2.840} $\pm$ \textbf{0.002} & \textbf{4.098} $\pm$ 0.040 & \textbf{4.853} $\pm$ \textbf{0.059} \\
          &   & SDS & 1.341 $\pm$ 0.012 & 1.858 $\pm$ 0.016 & 2.854 $\pm$ 0.003 & 4.107 $\pm$ 0.039 & \textbf{4.853} $\pm$ \textbf{0.057} \\
\newpage
        \multirow{16}{1cm}{Zoom Blur} & \multirow{4}{0.8cm}{Accuracy} & DS & 0.547 $\pm$ 0.005 & 0.448 $\pm$ 0.004 & \textbf{0.377} $\pm$ 0.003 & \textbf{0.311} $\pm$ 0.004 & \textbf{0.251} $\pm$ 0.004 \\
          &   & BCC & 0.024 $\pm$ 0.002 & 0.010 $\pm$ 0.002 & 0.001 $\pm$ 0.000 & 0.001 $\pm$ 0.000 & 0.001 $\pm$ 0.000 \\
          &   & EA & \textbf{0.550} $\pm$ 0.004 & \textbf{0.449} $\pm$ 0.004 & 0.376 $\pm$ 0.004 & 0.309 $\pm$ 0.005 & 0.247 $\pm$ 0.005 \\
          &   & SDS & \textbf{0.550} $\pm$ 0.004 & \textbf{0.449} $\pm$ 0.004 & 0.376 $\pm$ 0.004 & 0.309 $\pm$ 0.005 & 0.247 $\pm$ 0.005 \\
        \cmidrule{2-8}
          & \multirow{4}{0.8cm}{ECE} & DS & 0.446 $\pm$ 0.005 & 0.539 $\pm$ 0.004 & 0.605 $\pm$ 0.002 & 0.661 $\pm$ 0.004 & 0.707 $\pm$ 0.004 \\
          &   & BCC & 0.819 $\pm$ 0.005 & 0.849 $\pm$ 0.006 & 0.948 $\pm$ 0.002 & 0.943 $\pm$ 0.001 & 0.939 $\pm$ 0.005 \\
          &   & EA & \textbf{0.015} $\pm$ \textbf{0.001} & 0.018 $\pm$ 0.002 & 0.038 $\pm$ 0.001 & 0.062 $\pm$ 0.002 & 0.083 $\pm$ 0.002 \\
          &   & SDS & 0.020 $\pm$ 0.001 & \textbf{0.012} $\pm$ \textbf{0.001} & \textbf{0.030} $\pm$ \textbf{0.001} & \textbf{0.055} $\pm$ \textbf{0.002} & \textbf{0.077} $\pm$ \textbf{0.002} \\
        \cmidrule{2-8}
          & \multirow{4}{0.8cm}{Brier score} & DS & 0.896 $\pm$ 0.010 & 1.084 $\pm$ 0.007 & 1.219 $\pm$ 0.004 & 1.336 $\pm$ 0.009 & 1.436 $\pm$ 0.008 \\
          &   & BCC & 1.745 $\pm$ 0.006 & 1.792 $\pm$ 0.007 & 1.920 $\pm$ 0.002 & 1.915 $\pm$ 0.001 & 1.909 $\pm$ 0.007 \\
          &   & EA & \textbf{0.579} $\pm$ \textbf{0.004} & 0.688 $\pm$ 0.004 & \textbf{0.760} $\pm$ \textbf{0.004} & 0.825 $\pm$ 0.004 & 0.879 $\pm$ 0.003 \\
          &   & SDS & \textbf{0.579} $\pm$ \textbf{0.004} & \textbf{0.687} $\pm$ 0.004 & \textbf{0.760} $\pm$ \textbf{0.004} & \textbf{0.824} $\pm$ 0.004 & \textbf{0.877} $\pm$ 0.003 \\
        \cmidrule{2-8}
          & \multirow{4}{0.8cm}{NLL} & DS & 11.664 $\pm$ 0.148 & 14.234 $\pm$ 0.113 & 16.143 $\pm$ 0.039 & 17.893 $\pm$ 0.204 & 19.324 $\pm$ 0.158 \\
          &   & BCC & 10.036 $\pm$ 0.030 & 10.805 $\pm$ 0.036 & 12.585 $\pm$ 0.029 & 13.114 $\pm$ 0.055 & 13.717 $\pm$ 0.230 \\
          &   & EA & \textbf{2.028} $\pm$ 0.013 & \textbf{2.660} $\pm$ 0.016 & \textbf{3.118} $\pm$ 0.020 & \textbf{3.626} $\pm$ 0.023 & 4.125 $\pm$ 0.019 \\
          &   & SDS & 2.043 $\pm$ 0.011 & 2.674 $\pm$ 0.015 & 3.130 $\pm$ 0.024 & 3.628 $\pm$ 0.025 & \textbf{4.117} $\pm$ 0.021 \\
\end{longtable}
}

{
\scriptsize
\begin{longtable}{p{1cm}p{0.8cm}p{0.8cm}llllllll}
    \caption{Results on ImageNet with weather corruptions of increasing severity.}
    \label{tab:imagenet_weather}\\
        \toprule
        Corruption & Metric & Method & Severity 1 & Severity 2 & Severity 3 & Severity 4 & Severity 5 \\
        \midrule
        \endfirsthead
    \caption[]{Results on ImageNet with weather corruptions of increasing severity (continued).}\\
        \toprule
        Corruption & Metric & Method & Severity 1 & Severity 2 & Severity 3 & Severity 4 & Severity 5 \\
        \midrule
        \endhead
        \midrule
        \multicolumn{9}{r}{\textit{Continued on next page}}\\
        \endfoot
        \bottomrule
        \endlastfoot
        \multirow{16}{1cm}{Snow} & \multirow{4}{0.8cm}{Accuracy} & DS & 0.557 $\pm$ 0.001 & \textbf{0.321} $\pm$ 0.001 & \textbf{0.360} $\pm$ 0.001 & \textbf{0.251} $\pm$ \textbf{0.001} & \textbf{0.176} $\pm$ 0.002 \\
          &   & BCC & 0.091 $\pm$ 0.022 & 0.006 $\pm$ 0.003 & 0.008 $\pm$ 0.006 & 0.001 $\pm$ 0.000 & 0.003 $\pm$ 0.001 \\
          &   & EA & 0.560 $\pm$ 0.001 & 0.319 $\pm$ 0.002 & 0.359 $\pm$ 0.002 & 0.248 $\pm$ 0.002 & 0.173 $\pm$ 0.002 \\
          &   & SDS & \textbf{0.561} $\pm$ 0.001 & 0.319 $\pm$ 0.002 & 0.359 $\pm$ 0.002 & 0.248 $\pm$ 0.002 & 0.173 $\pm$ 0.002 \\
        \cmidrule{2-8}
          & \multirow{4}{0.8cm}{ECE} & DS & 0.435 $\pm$ 0.001 & 0.643 $\pm$ 0.001 & 0.612 $\pm$ 0.001 & 0.695 $\pm$ 0.003 & 0.743 $\pm$ 0.004 \\
          &   & BCC & 0.772 $\pm$ 0.009 & 0.893 $\pm$ 0.032 & 0.883 $\pm$ 0.045 & 0.922 $\pm$ 0.011 & 0.914 $\pm$ 0.009 \\
          &   & EA & \textbf{0.016} $\pm$ 0.002 & 0.068 $\pm$ 0.006 & 0.052 $\pm$ 0.006 & 0.081 $\pm$ 0.007 & 0.137 $\pm$ 0.006 \\
          &   & SDS & 0.017 $\pm$ 0.002 & \textbf{0.061} $\pm$ 0.006 & \textbf{0.044} $\pm$ 0.006 & \textbf{0.075} $\pm$ 0.007 & \textbf{0.131} $\pm$ 0.006 \\
        \cmidrule{2-8}
          & \multirow{4}{0.8cm}{Brier score} & DS & 0.875 $\pm$ 0.002 & 1.305 $\pm$ 0.002 & 1.238 $\pm$ 0.002 & 1.420 $\pm$ 0.004 & 1.531 $\pm$ 0.005 \\
          &   & BCC & 1.640 $\pm$ 0.024 & 1.852 $\pm$ 0.040 & 1.838 $\pm$ 0.058 & 1.886 $\pm$ 0.017 & 1.882 $\pm$ 0.008 \\
          &   & EA & \textbf{0.565} $\pm$ \textbf{0.002} & 0.814 $\pm$ 0.004 & 0.775 $\pm$ 0.003 & 0.876 $\pm$ 0.003 & 0.962 $\pm$ 0.006 \\
          &   & SDS & \textbf{0.565} $\pm$ \textbf{0.001} & \textbf{0.813} $\pm$ 0.003 & \textbf{0.774} $\pm$ 0.003 & \textbf{0.874} $\pm$ 0.003 & \textbf{0.959} $\pm$ 0.006 \\
        \cmidrule{2-8}
          & \multirow{4}{0.8cm}{NLL} & DS & 11.418 $\pm$ 0.054 & 17.489 $\pm$ 0.066 & 16.593 $\pm$ 0.006 & 19.328 $\pm$ 0.134 & 20.635 $\pm$ 0.047 \\
          &   & BCC & 9.710 $\pm$ 0.087 & 12.396 $\pm$ 0.611 & 11.998 $\pm$ 0.634 & 13.809 $\pm$ 0.262 & 14.700 $\pm$ 0.276 \\
          &   & EA & \textbf{1.981} $\pm$ 0.008 & \textbf{3.619} $\pm$ 0.035 & \textbf{3.355} $\pm$ 0.030 & \textbf{4.234} $\pm$ 0.041 & 4.935 $\pm$ 0.032 \\
          &   & SDS & 1.994 $\pm$ 0.006 & 3.631 $\pm$ 0.034 & 3.369 $\pm$ 0.031 & 4.244 $\pm$ 0.041 & \textbf{4.924} $\pm$ 0.033 \\
        \midrule
        \multirow{16}{1cm}{Frost} & \multirow{4}{0.8cm}{Accuracy} & DS & 0.625 $\pm$ 0.001 & 0.456 $\pm$ 0.001 & \textbf{0.338} $\pm$ 0.002 & \textbf{0.316} $\pm$ 0.003 & \textbf{0.246} $\pm$ 0.003 \\
          &   & BCC & 0.608 $\pm$ 0.001 & 0.023 $\pm$ 0.018 & 0.007 $\pm$ 0.008 & 0.002 $\pm$ 0.001 & 0.002 $\pm$ 0.001 \\
          &   & EA & 0.631 $\pm$ 0.001 & \textbf{0.458} $\pm$ 0.002 & 0.337 $\pm$ 0.001 & 0.314 $\pm$ 0.002 & 0.244 $\pm$ 0.003 \\
          &   & SDS & \textbf{0.632} $\pm$ 0.001 & \textbf{0.458} $\pm$ 0.002 & 0.337 $\pm$ 0.001 & 0.314 $\pm$ 0.002 & 0.244 $\pm$ 0.003 \\
        \cmidrule{2-8}
          & \multirow{4}{0.8cm}{ECE} & DS & 0.371 $\pm$ 0.001 & 0.533 $\pm$ 0.001 & 0.639 $\pm$ 0.002 & 0.656 $\pm$ 0.004 & 0.710 $\pm$ 0.003 \\
          &   & BCC & 0.607 $\pm$ 0.001 & 0.858 $\pm$ 0.065 & 0.900 $\pm$ 0.046 & 0.926 $\pm$ 0.005 & 0.915 $\pm$ 0.007 \\
          &   & EA & \textbf{0.023} $\pm$ \textbf{0.000} & \textbf{0.015} $\pm$ \textbf{0.001} & 0.041 $\pm$ 0.003 & 0.052 $\pm$ 0.003 & 0.077 $\pm$ 0.002 \\
          &   & SDS & 0.027 $\pm$ 0.000 & \textbf{0.015} $\pm$ \textbf{0.002} & \textbf{0.034} $\pm$ \textbf{0.003} & \textbf{0.046} $\pm$ 0.003 & \textbf{0.071} $\pm$ \textbf{0.002} \\
        \cmidrule{2-8}
          & \multirow{4}{0.8cm}{Brier score} & DS & 0.744 $\pm$ 0.002 & 1.071 $\pm$ 0.002 & 1.289 $\pm$ 0.004 & 1.326 $\pm$ 0.006 & 1.443 $\pm$ 0.006 \\
          &   & BCC & 0.999 $\pm$ 0.000 & 1.794 $\pm$ 0.090 & 1.859 $\pm$ 0.059 & 1.892 $\pm$ 0.005 & 1.878 $\pm$ 0.007 \\
          &   & EA & \textbf{0.486} $\pm$ \textbf{0.002} & \textbf{0.671} $\pm$ 0.001 & 0.794 $\pm$ 0.002 & 0.817 $\pm$ 0.002 & 0.882 $\pm$ 0.002 \\
          &   & SDS & \textbf{0.486} $\pm$ \textbf{0.002} & 0.672 $\pm$ 0.001 & \textbf{0.793} $\pm$ 0.002 & \textbf{0.816} $\pm$ 0.002 & \textbf{0.880} $\pm$ 0.002 \\
        \cmidrule{2-8}
          & \multirow{4}{0.8cm}{NLL} & DS & 9.604 $\pm$ 0.050 & 14.216 $\pm$ 0.047 & 17.375 $\pm$ 0.068 & 17.914 $\pm$ 0.102 & 19.686 $\pm$ 0.033 \\
          &   & BCC & 6.907 $\pm$ 0.000 & 10.989 $\pm$ 1.037 & 12.268 $\pm$ 0.750 & 13.004 $\pm$ 0.277 & 13.533 $\pm$ 0.307 \\
          &   & EA & \textbf{1.591} $\pm$ 0.006 & \textbf{2.613} $\pm$ \textbf{0.006} & \textbf{3.478} $\pm$ \textbf{0.003} & \textbf{3.678} $\pm$ \textbf{0.004} & \textbf{4.269} $\pm$ \textbf{0.002} \\
          &   & SDS & 1.599 $\pm$ 0.007 & 2.628 $\pm$ 0.006 & 3.492 $\pm$ 0.003 & 3.691 $\pm$ 0.003 & 4.276 $\pm$ 0.002 \\
\newpage
        \multirow{16}{1cm}{Fog} & \multirow{4}{0.8cm}{Accuracy} & DS & 0.628 $\pm$ 0.004 & 0.569 $\pm$ 0.004 & 0.478 $\pm$ 0.006 & 0.409 $\pm$ 0.004 & \textbf{0.245} $\pm$ 0.004 \\
          &   & BCC & 0.608 $\pm$ 0.005 & 0.001 $\pm$ 0.000 & 0.027 $\pm$ 0.002 & 0.020 $\pm$ 0.002 & 0.003 $\pm$ 0.000 \\
          &   & EA & 0.633 $\pm$ 0.005 & \textbf{0.572} $\pm$ 0.005 & \textbf{0.481} $\pm$ 0.006 & 0.409 $\pm$ 0.005 & 0.242 $\pm$ 0.004 \\
          &   & SDS & \textbf{0.634} $\pm$ 0.005 & \textbf{0.572} $\pm$ 0.006 & \textbf{0.481} $\pm$ 0.006 & 0.409 $\pm$ 0.005 & 0.242 $\pm$ 0.004 \\
        \cmidrule{2-8}
          & \multirow{4}{0.8cm}{ECE} & DS & 0.368 $\pm$ 0.004 & 0.425 $\pm$ 0.004 & 0.510 $\pm$ 0.005 & 0.569 $\pm$ 0.003 & 0.698 $\pm$ 0.004 \\
          &   & BCC & 0.607 $\pm$ 0.005 & 0.000 $\pm$ 0.000 & 0.827 $\pm$ 0.003 & 0.834 $\pm$ 0.003 & 0.907 $\pm$ 0.003 \\
          &   & EA & \textbf{0.021} $\pm$ \textbf{0.000} & \textbf{0.016} $\pm$ \textbf{0.002} & 0.013 $\pm$ 0.002 & 0.040 $\pm$ 0.003 & 0.114 $\pm$ 0.005 \\
          &   & SDS & 0.025 $\pm$ 0.001 & 0.020 $\pm$ 0.002 & \textbf{0.008} $\pm$ \textbf{0.000} & \textbf{0.032} $\pm$ \textbf{0.003} & \textbf{0.107} $\pm$ 0.005 \\
        \cmidrule{2-8}
          & \multirow{4}{0.8cm}{Brier score} & DS & 0.737 $\pm$ 0.008 & 0.853 $\pm$ 0.008 & 1.026 $\pm$ 0.011 & 1.148 $\pm$ 0.007 & 1.425 $\pm$ 0.007 \\
          &   & BCC & 0.999 $\pm$ 0.000 & 0.999 $\pm$ 0.000 & 1.752 $\pm$ 0.004 & 1.770 $\pm$ 0.004 & 1.868 $\pm$ 0.004 \\
          &   & EA & \textbf{0.487} $\pm$ \textbf{0.005} & \textbf{0.556} $\pm$ \textbf{0.006} & \textbf{0.656} $\pm$ \textbf{0.006} & \textbf{0.731} $\pm$ \textbf{0.005} & 0.899 $\pm$ 0.006 \\
          &   & SDS & \textbf{0.487} $\pm$ \textbf{0.005} & \textbf{0.556} $\pm$ \textbf{0.006} & \textbf{0.656} $\pm$ \textbf{0.006} & \textbf{0.731} $\pm$ \textbf{0.005} & \textbf{0.897} $\pm$ 0.006 \\
        \cmidrule{2-8}
          & \multirow{4}{0.8cm}{NLL} & DS & 9.442 $\pm$ 0.095 & 10.995 $\pm$ 0.103 & 13.373 $\pm$ 0.144 & 14.985 $\pm$ 0.074 & 18.564 $\pm$ 0.162 \\
          &   & BCC & 6.907 $\pm$ 0.000 & 6.907 $\pm$ 0.000 & 10.455 $\pm$ 0.036 & 11.251 $\pm$ 0.074 & 14.556 $\pm$ 0.180 \\
          &   & EA & \textbf{1.529} $\pm$ 0.020 & \textbf{1.844} $\pm$ 0.028 & \textbf{2.364} $\pm$ 0.033 & \textbf{2.841} $\pm$ 0.027 & \textbf{4.168} $\pm$ 0.036 \\
          &   & SDS & 1.538 $\pm$ 0.021 & 1.856 $\pm$ 0.032 & 2.379 $\pm$ 0.034 & 2.855 $\pm$ 0.028 & 4.175 $\pm$ 0.037 \\
        \midrule
        \multirow{16}{1cm}{Brightness} & \multirow{4}{0.8cm}{Accuracy} & DS & 0.754 $\pm$ 0.002 & 0.737 $\pm$ 0.002 & 0.711 $\pm$ 0.002 & 0.668 $\pm$ 0.001 & 0.606 $\pm$ 0.002 \\
          &   & BCC & 0.748 $\pm$ 0.002 & 0.730 $\pm$ 0.002 & 0.702 $\pm$ 0.002 & 0.655 $\pm$ 0.001 & 0.582 $\pm$ 0.002 \\
          &   & EA & \textbf{0.759} $\pm$ \textbf{0.002} & \textbf{0.743} $\pm$ \textbf{0.002} & 0.716 $\pm$ 0.002 & 0.673 $\pm$ 0.001 & \textbf{0.611} $\pm$ \textbf{0.001} \\
          &   & SDS & \textbf{0.759} $\pm$ \textbf{0.002} & \textbf{0.743} $\pm$ \textbf{0.002} & \textbf{0.717} $\pm$ 0.002 & \textbf{0.674} $\pm$ 0.001 & \textbf{0.611} $\pm$ \textbf{0.001} \\
        \cmidrule{2-8}
          & \multirow{4}{0.8cm}{ECE} & DS & 0.244 $\pm$ 0.002 & 0.260 $\pm$ 0.002 & 0.286 $\pm$ 0.002 & 0.328 $\pm$ 0.001 & 0.389 $\pm$ 0.002 \\
          &   & BCC & 0.747 $\pm$ 0.002 & 0.729 $\pm$ 0.002 & 0.701 $\pm$ 0.002 & 0.654 $\pm$ 0.001 & 0.581 $\pm$ 0.002 \\
          &   & EA & \textbf{0.021} $\pm$ \textbf{0.001} & \textbf{0.023} $\pm$ \textbf{0.001} & \textbf{0.026} $\pm$ \textbf{0.000} & \textbf{0.028} $\pm$ \textbf{0.001} & \textbf{0.027} $\pm$ \textbf{0.001} \\
          &   & SDS & \textbf{0.021} $\pm$ \textbf{0.001} & \textbf{0.023} $\pm$ \textbf{0.001} & 0.028 $\pm$ 0.001 & 0.032 $\pm$ 0.001 & 0.033 $\pm$ 0.001 \\
        \cmidrule{2-8}
          & \multirow{4}{0.8cm}{Brier score} & DS & 0.489 $\pm$ 0.004 & 0.522 $\pm$ 0.004 & 0.574 $\pm$ 0.005 & 0.659 $\pm$ 0.002 & 0.781 $\pm$ 0.004 \\
          &   & BCC & 0.999 $\pm$ 0.000 & 0.999 $\pm$ 0.000 & 0.999 $\pm$ 0.000 & 0.999 $\pm$ 0.000 & 0.999 $\pm$ 0.000 \\
          &   & EA & \textbf{0.336} $\pm$ \textbf{0.002} & \textbf{0.357} $\pm$ \textbf{0.002} & \textbf{0.389} $\pm$ \textbf{0.002} & \textbf{0.441} $\pm$ \textbf{0.001} & \textbf{0.514} $\pm$ \textbf{0.001} \\
          &   & SDS & \textbf{0.336} $\pm$ \textbf{0.002} & \textbf{0.357} $\pm$ \textbf{0.002} & \textbf{0.389} $\pm$ \textbf{0.002} & \textbf{0.441} $\pm$ \textbf{0.001} & \textbf{0.514} $\pm$ \textbf{0.001} \\
        \cmidrule{2-8}
          & \multirow{4}{0.8cm}{NLL} & DS & 6.130 $\pm$ 0.059 & 6.554 $\pm$ 0.045 & 7.236 $\pm$ 0.057 & 8.335 $\pm$ 0.029 & 9.979 $\pm$ 0.057 \\
          &   & BCC & 6.907 $\pm$ 0.000 & 6.907 $\pm$ 0.000 & 6.907 $\pm$ 0.000 & 6.907 $\pm$ 0.000 & 6.907 $\pm$ 0.000 \\
          &   & EA & \textbf{0.947} $\pm$ 0.006 & \textbf{1.022} $\pm$ 0.007 & \textbf{1.144} $\pm$ 0.007 & \textbf{1.347} $\pm$ 0.007 & \textbf{1.657} $\pm$ \textbf{0.006} \\
          &   & SDS & 0.956 $\pm$ 0.006 & 1.032 $\pm$ 0.007 & 1.154 $\pm$ 0.008 & 1.358 $\pm$ 0.007 & 1.669 $\pm$ 0.006 \\
\end{longtable}
}

{
\scriptsize
\begin{longtable}{p{1cm}p{0.8cm}p{0.8cm}llllllll}
    \caption{Results on ImageNet with digital corruptions of increasing severity.}
    \label{tab:imagenet_digital}\\
        \toprule
        Corruption & Metric & Method & Severity 1 & Severity 2 & Severity 3 & Severity 4 & Severity 5 \\
        \midrule
        \endfirsthead
    \caption[]{Results on ImageNet with digital corruptions of increasing severity (continued).}\\
        \toprule
        Corruption & Metric & Method & Severity 1 & Severity 2 & Severity 3 & Severity 4 & Severity 5 \\
        \midrule
        \endhead
        \midrule
        \multicolumn{9}{r}{\textit{Continued on next page}}\\
        \endfoot
        \bottomrule
        \endlastfoot
        \multirow{16}{1cm}{Contrast} & \multirow{4}{0.8cm}{Accuracy} & DS & 0.667 $\pm$ 0.002 & 0.608 $\pm$ 0.003 & 0.492 $\pm$ 0.004 & \textbf{0.229} $\pm$ 0.011 & \textbf{0.055} $\pm$ 0.004 \\
          &   & BCC & 0.654 $\pm$ 0.002 & 0.001 $\pm$ 0.000 & 0.015 $\pm$ 0.001 & 0.004 $\pm$ 0.000 & 0.002 $\pm$ 0.000 \\
          &   & EA & \textbf{0.673} $\pm$ \textbf{0.001} & \textbf{0.613} $\pm$ 0.003 & \textbf{0.494} $\pm$ 0.004 & 0.227 $\pm$ 0.010 & 0.054 $\pm$ 0.003 \\
          &   & SDS & \textbf{0.673} $\pm$ \textbf{0.001} & \textbf{0.613} $\pm$ 0.003 & \textbf{0.494} $\pm$ 0.004 & 0.227 $\pm$ 0.010 & 0.054 $\pm$ 0.003 \\
        \cmidrule{2-8}
          & \multirow{4}{0.8cm}{ECE} & DS & 0.329 $\pm$ 0.002 & 0.387 $\pm$ 0.003 & 0.497 $\pm$ 0.003 & 0.717 $\pm$ 0.007 & 0.752 $\pm$ 0.005 \\
          &   & BCC & 0.653 $\pm$ 0.002 & 0.000 $\pm$ 0.000 & 0.852 $\pm$ 0.002 & 0.915 $\pm$ 0.002 & 0.930 $\pm$ 0.015 \\
          &   & EA & \textbf{0.022} $\pm$ \textbf{0.000} & \textbf{0.023} $\pm$ \textbf{0.000} & \textbf{0.016} $\pm$ \textbf{0.001} & 0.050 $\pm$ 0.006 & 0.072 $\pm$ 0.003 \\
          &   & SDS & 0.028 $\pm$ 0.000 & 0.028 $\pm$ 0.001 & 0.019 $\pm$ 0.001 & \textbf{0.045} $\pm$ 0.007 & \textbf{0.069} $\pm$ 0.003 \\
        \cmidrule{2-8}
          & \multirow{4}{0.8cm}{Brier score} & DS & 0.660 $\pm$ 0.004 & 0.776 $\pm$ 0.006 & 0.999 $\pm$ 0.007 & 1.463 $\pm$ 0.016 & 1.622 $\pm$ 0.007 \\
          &   & BCC & 0.999 $\pm$ 0.000 & 0.999 $\pm$ 0.000 & 1.788 $\pm$ 0.002 & 1.881 $\pm$ 0.003 & 1.897 $\pm$ 0.023 \\
          &   & EA & \textbf{0.441} $\pm$ \textbf{0.002} & \textbf{0.512} $\pm$ \textbf{0.003} & \textbf{0.643} $\pm$ \textbf{0.005} & 0.892 $\pm$ 0.008 & 0.997 $\pm$ 0.003 \\
          &   & SDS & \textbf{0.441} $\pm$ \textbf{0.002} & \textbf{0.512} $\pm$ \textbf{0.003} & \textbf{0.643} $\pm$ \textbf{0.005} & \textbf{0.891} $\pm$ 0.008 & \textbf{0.996} $\pm$ 0.003 \\
        \cmidrule{2-8}
          & \multirow{4}{0.8cm}{NLL} & DS & 8.359 $\pm$ 0.065 & 9.883 $\pm$ 0.101 & 12.918 $\pm$ 0.034 & 19.387 $\pm$ 0.267 & 23.885 $\pm$ 0.113 \\
          &   & BCC & 6.907 $\pm$ 0.000 & 6.907 $\pm$ 0.000 & 10.459 $\pm$ 0.043 & 12.987 $\pm$ 0.209 & 16.973 $\pm$ 0.466 \\
          &   & EA & \textbf{1.333} $\pm$ 0.009 & \textbf{1.631} $\pm$ 0.014 & \textbf{2.281} $\pm$ 0.032 & \textbf{4.150} $\pm$ 0.087 & 6.154 $\pm$ 0.093 \\
          &   & SDS & 1.348 $\pm$ 0.009 & 1.649 $\pm$ 0.014 & 2.297 $\pm$ 0.032 & 4.153 $\pm$ 0.085 & \textbf{6.107} $\pm$ 0.088 \\
\newpage
        \multirow{16}{1cm}{Elastic} & \multirow{4}{0.8cm}{Accuracy} & DS & 0.688 $\pm$ 0.003 & 0.473 $\pm$ 0.003 & 0.584 $\pm$ 0.004 & \textbf{0.455} $\pm$ 0.004 & \textbf{0.207} $\pm$ 0.005 \\
          &   & BCC & 0.679 $\pm$ 0.003 & 0.022 $\pm$ 0.034 & 0.001 $\pm$ 0.000 & 0.025 $\pm$ 0.003 & 0.003 $\pm$ 0.000 \\
          &   & EA & \textbf{0.694} $\pm$ \textbf{0.002} & \textbf{0.475} $\pm$ 0.002 & 0.584 $\pm$ 0.004 & 0.452 $\pm$ 0.006 & 0.200 $\pm$ 0.006 \\
          &   & SDS & \textbf{0.694} $\pm$ \textbf{0.002} & \textbf{0.475} $\pm$ 0.002 & 0.584 $\pm$ 0.004 & 0.452 $\pm$ 0.006 & 0.200 $\pm$ 0.006 \\
        \cmidrule{2-8}
          & \multirow{4}{0.8cm}{ECE} & DS & 0.309 $\pm$ 0.003 & 0.517 $\pm$ 0.003 & 0.410 $\pm$ 0.004 & 0.530 $\pm$ 0.004 & 0.707 $\pm$ 0.012 \\
          &   & BCC & 0.678 $\pm$ 0.003 & 0.883 $\pm$ 0.102 & 0.423 $\pm$ 0.196 & 0.850 $\pm$ 0.014 & 0.929 $\pm$ 0.004 \\
          &   & EA & \textbf{0.030} $\pm$ \textbf{0.000} & 0.021 $\pm$ 0.001 & \textbf{0.014} $\pm$ 0.002 & 0.033 $\pm$ 0.002 & 0.168 $\pm$ 0.004 \\
          &   & SDS & 0.036 $\pm$ 0.000 & \textbf{0.020} $\pm$ 0.001 & 0.017 $\pm$ 0.002 & \textbf{0.024} $\pm$ \textbf{0.002} & \textbf{0.161} $\pm$ 0.004 \\
        \cmidrule{2-8}
          & \multirow{4}{0.8cm}{Brier score} & DS & 0.619 $\pm$ 0.005 & 1.039 $\pm$ 0.006 & 0.823 $\pm$ 0.007 & 1.067 $\pm$ 0.008 & 1.462 $\pm$ 0.018 \\
          &   & BCC & 0.999 $\pm$ 0.000 & 1.829 $\pm$ 0.145 & 1.378 $\pm$ 0.202 & 1.786 $\pm$ 0.017 & 1.904 $\pm$ 0.005 \\
          &   & EA & \textbf{0.415} $\pm$ \textbf{0.002} & \textbf{0.654} $\pm$ \textbf{0.002} & \textbf{0.540} $\pm$ \textbf{0.003} & \textbf{0.683} $\pm$ \textbf{0.004} & 0.963 $\pm$ 0.004 \\
          &   & SDS & \textbf{0.415} $\pm$ \textbf{0.002} & \textbf{0.654} $\pm$ \textbf{0.002} & \textbf{0.540} $\pm$ \textbf{0.003} & \textbf{0.683} $\pm$ \textbf{0.004} & \textbf{0.959} $\pm$ 0.004 \\
        \cmidrule{2-8}
          & \multirow{4}{0.8cm}{NLL} & DS & 7.859 $\pm$ 0.050 & 13.785 $\pm$ 0.097 & 10.591 $\pm$ 0.115 & 13.934 $\pm$ 0.172 & 19.179 $\pm$ 0.327 \\
          &   & BCC & 6.907 $\pm$ 0.000 & 11.302 $\pm$ 1.044 & 8.692 $\pm$ 0.922 & 11.037 $\pm$ 0.082 & 14.775 $\pm$ 0.104 \\
          &   & EA & \textbf{1.259} $\pm$ 0.009 & \textbf{2.643} $\pm$ 0.013 & \textbf{1.877} $\pm$ 0.017 & \textbf{2.705} $\pm$ 0.032 & 4.796 $\pm$ 0.039 \\
          &   & SDS & 1.270 $\pm$ 0.009 & 2.658 $\pm$ 0.013 & 1.892 $\pm$ 0.017 & 2.707 $\pm$ 0.031 & \textbf{4.737} $\pm$ 0.039 \\
        \midrule
        \multirow{16}{1cm}{Pixelate} & \multirow{4}{0.8cm}{Accuracy} & DS & 0.660 $\pm$ 0.005 & 0.658 $\pm$ 0.007 & 0.520 $\pm$ 0.009 & 0.372 $\pm$ 0.029 & 0.307 $\pm$ 0.045 \\
          &   & BCC & 0.640 $\pm$ 0.005 & 0.636 $\pm$ 0.007 & 0.013 $\pm$ 0.010 & 0.003 $\pm$ 0.001 & 0.002 $\pm$ 0.001 \\
          &   & EA & \textbf{0.665} $\pm$ 0.004 & 0.663 $\pm$ 0.007 & \textbf{0.525} $\pm$ 0.010 & \textbf{0.375} $\pm$ 0.028 & \textbf{0.309} $\pm$ 0.043 \\
          &   & SDS & \textbf{0.665} $\pm$ 0.004 & \textbf{0.664} $\pm$ 0.007 & \textbf{0.525} $\pm$ 0.010 & \textbf{0.375} $\pm$ 0.028 & \textbf{0.309} $\pm$ 0.043 \\
        \cmidrule{2-8}
          & \multirow{4}{0.8cm}{ECE} & DS & 0.336 $\pm$ 0.005 & 0.337 $\pm$ 0.007 & 0.470 $\pm$ 0.008 & 0.599 $\pm$ 0.021 & 0.648 $\pm$ 0.027 \\
          &   & BCC & 0.639 $\pm$ 0.005 & 0.635 $\pm$ 0.007 & 0.888 $\pm$ 0.049 & 0.915 $\pm$ 0.030 & 0.929 $\pm$ 0.019 \\
          &   & EA & \textbf{0.018} $\pm$ 0.002 & \textbf{0.018} $\pm$ 0.002 & \textbf{0.020} $\pm$ \textbf{0.003} & 0.031 $\pm$ 0.006 & 0.041 $\pm$ 0.016 \\
          &   & SDS & 0.019 $\pm$ 0.003 & 0.019 $\pm$ 0.003 & \textbf{0.020} $\pm$ \textbf{0.006} & \textbf{0.028} $\pm$ 0.002 & \textbf{0.036} $\pm$ 0.014 \\
        \cmidrule{2-8}
          & \multirow{4}{0.8cm}{Brier score} & DS & 0.673 $\pm$ 0.010 & 0.676 $\pm$ 0.013 & 0.944 $\pm$ 0.017 & 1.214 $\pm$ 0.047 & 1.322 $\pm$ 0.064 \\
          &   & BCC & 0.999 $\pm$ 0.000 & 0.999 $\pm$ 0.000 & 1.834 $\pm$ 0.070 & 1.876 $\pm$ 0.038 & 1.895 $\pm$ 0.026 \\
          &   & EA & \textbf{0.452} $\pm$ \textbf{0.005} & \textbf{0.454} $\pm$ \textbf{0.008} & \textbf{0.611} $\pm$ \textbf{0.010} & \textbf{0.760} $\pm$ \textbf{0.024} & \textbf{0.820} $\pm$ \textbf{0.036} \\
          &   & SDS & \textbf{0.452} $\pm$ \textbf{0.005} & \textbf{0.454} $\pm$ \textbf{0.008} & \textbf{0.611} $\pm$ \textbf{0.010} & \textbf{0.760} $\pm$ \textbf{0.024} & \textbf{0.820} $\pm$ \textbf{0.036} \\
        \cmidrule{2-8}
          & \multirow{4}{0.8cm}{NLL} & DS & 8.520 $\pm$ 0.126 & 8.553 $\pm$ 0.191 & 12.217 $\pm$ 0.240 & 15.971 $\pm$ 0.622 & 17.632 $\pm$ 0.959 \\
          &   & BCC & 6.907 $\pm$ 0.000 & 6.907 $\pm$ 0.000 & 10.794 $\pm$ 0.822 & 11.912 $\pm$ 0.650 & 12.871 $\pm$ 1.027 \\
          &   & EA & \textbf{1.388} $\pm$ 0.018 & \textbf{1.392} $\pm$ 0.030 & \textbf{2.124} $\pm$ 0.055 & \textbf{3.059} $\pm$ 0.184 & \textbf{3.551} $\pm$ 0.316 \\
          &   & SDS & 1.401 $\pm$ 0.017 & 1.402 $\pm$ 0.030 & 2.141 $\pm$ 0.055 & 3.070 $\pm$ 0.180 & 3.563 $\pm$ 0.312 \\
        \midrule
        \multirow{16}{1cm}{JPEG} & \multirow{4}{0.8cm}{Accuracy} & DS & 0.680 $\pm$ 0.002 & 0.646 $\pm$ 0.004 & 0.620 $\pm$ 0.007 & 0.529 $\pm$ 0.017 & 0.403 $\pm$ 0.031 \\
          &   & BCC & 0.666 $\pm$ 0.002 & 0.627 $\pm$ 0.007 & 0.551 $\pm$ 0.067 & 0.001 $\pm$ 0.000 & 0.004 $\pm$ 0.005 \\
          &   & EA & \textbf{0.684} $\pm$ \textbf{0.002} & \textbf{0.650} $\pm$ 0.004 & \textbf{0.625} $\pm$ 0.007 & \textbf{0.533} $\pm$ 0.016 & \textbf{0.405} $\pm$ 0.030 \\
          &   & SDS & \textbf{0.684} $\pm$ 0.003 & \textbf{0.650} $\pm$ 0.004 & \textbf{0.625} $\pm$ 0.007 & \textbf{0.533} $\pm$ 0.016 & \textbf{0.405} $\pm$ 0.030 \\
        \cmidrule{2-8}
          & \multirow{4}{0.8cm}{ECE} & DS & 0.316 $\pm$ 0.003 & 0.349 $\pm$ 0.004 & 0.374 $\pm$ 0.007 & 0.462 $\pm$ 0.016 & 0.578 $\pm$ 0.025 \\
          &   & BCC & 0.665 $\pm$ 0.002 & 0.626 $\pm$ 0.007 & 0.550 $\pm$ 0.067 & 0.312 $\pm$ 0.540 & 0.923 $\pm$ 0.036 \\
          &   & EA & \textbf{0.019} $\pm$ 0.002 & \textbf{0.019} $\pm$ 0.001 & \textbf{0.020} $\pm$ 0.003 & \textbf{0.019} $\pm$ 0.005 & \textbf{0.021} $\pm$ 0.002 \\
          &   & SDS & 0.020 $\pm$ 0.002 & 0.021 $\pm$ 0.002 & 0.022 $\pm$ 0.004 & 0.022 $\pm$ 0.006 & 0.022 $\pm$ 0.003 \\
        \cmidrule{2-8}
          & \multirow{4}{0.8cm}{Brier score} & DS & 0.633 $\pm$ 0.005 & 0.700 $\pm$ 0.007 & 0.751 $\pm$ 0.014 & 0.928 $\pm$ 0.032 & 1.166 $\pm$ 0.054 \\
          &   & BCC & 0.999 $\pm$ 0.000 & 0.999 $\pm$ 0.000 & 0.999 $\pm$ 0.000 & 1.300 $\pm$ 0.521 & 1.885 $\pm$ 0.049 \\
          &   & EA & \textbf{0.428} $\pm$ 0.003 & \textbf{0.467} $\pm$ 0.005 & \textbf{0.497} $\pm$ 0.007 & \textbf{0.598} $\pm$ \textbf{0.018} & \textbf{0.728} $\pm$ \textbf{0.028} \\
          &   & SDS & 0.429 $\pm$ 0.003 & 0.468 $\pm$ 0.005 & 0.498 $\pm$ 0.007 & \textbf{0.598} $\pm$ \textbf{0.018} & \textbf{0.728} $\pm$ \textbf{0.028} \\
        \cmidrule{2-8}
          & \multirow{4}{0.8cm}{NLL} & DS & 7.978 $\pm$ 0.066 & 8.875 $\pm$ 0.054 & 9.549 $\pm$ 0.152 & 12.046 $\pm$ 0.435 & 15.377 $\pm$ 0.706 \\
          &   & BCC & 6.907 $\pm$ 0.000 & 6.907 $\pm$ 0.000 & 6.907 $\pm$ 0.000 & 8.473 $\pm$ 2.711 & 11.859 $\pm$ 0.860 \\
          &   & EA & \textbf{1.293} $\pm$ 0.011 & \textbf{1.456} $\pm$ 0.021 & \textbf{1.594} $\pm$ 0.034 & \textbf{2.103} $\pm$ 0.103 & \textbf{2.912} $\pm$ 0.210 \\
          &   & SDS & 1.309 $\pm$ 0.012 & 1.474 $\pm$ 0.021 & 1.612 $\pm$ 0.035 & 2.119 $\pm$ 0.102 & 2.927 $\pm$ 0.208 \\
\end{longtable}
}

\bibliography{references}

\end{document}